\documentclass[lettersize,journal]{IEEEtran}
\usepackage{amsmath,amsfonts}
\usepackage{algorithmic}
\usepackage{algorithm}
\usepackage{array}
\usepackage[caption=false,font=normalsize,labelfont=sf,textfont=sf]{subfig}
\usepackage{textcomp}
\usepackage{stfloats}
\usepackage{url}
\usepackage{verbatim}
\usepackage{graphicx}
\usepackage{cite}
\usepackage[export]{adjustbox}
\newcommand{\expect}{\mathbb{E}}
\usepackage{amssymb}
\usepackage{hyperref}
\usepackage{amsmath}
\usepackage{algorithmic}
\usepackage{siunitx}
\usepackage{breakurl}
\usepackage{xcolor}         
\usepackage{color}
\newcommand{\tp}{\mathsf{T}}

\begin{document}

\title{Spatially scalable recursive estimation of Gaussian process terrain maps using local basis functions}

\author{Frida Viset, Rudy Helmons, Manon Kok
\thanks{Frida Viset and Manon Kok are with the Department for Systems and Control, and Rudy Helmons is with the Marine and Transport Technology Department at the Delft University of Technology, Mekelweg 5, 2628 CD Delft, the Netherlands, (e-mail:{frida.viset@gmail.com, r.j.l.helmons@tudelft.nl, m.kok-1}@tudelft.nl). This publication is part of the project ``Sensor Fusion For Indoor localization Using The Magnetic Field'' with project number 18213 of the research program Veni which is (partly) financed by the Dutch Research Council (NWO). }
\thanks{}}

\markboth{Journal of \LaTeX\ Class Files,~Vol.~14, No.~8, August~2021}%
{Shell \MakeLowercase{\textit{et al.}}: A Sample Article Using IEEEtran.cls for IEEE Journals}

\IEEEpubid{}

\maketitle


\begin{abstract}
We address the computational challenges of large-scale geospatial mapping with Gaussian process (GP) regression by performing localized computations rather than processing the entire map simultaneously. Traditional approaches to GP regression often involve computational and storage costs that either scale with the number of measurements, or with the spatial extent of the mapped area, limiting their scalability for real-time applications. Our method places a global grid of finite-support basis functions and restricts computations to a local subset of the grid 1) surrounding the measurement when the map is updated, and 2) surrounding the query point when the map is queried. This localized approach ensures that only the relevant area is updated or queried at each timestep, significantly reducing computational complexity while maintaining accuracy. Unlike many existing methods, which suffer from boundary effects or increased computational costs with mapped area, our localized approach avoids discontinuities and ensures that computational costs remain manageable regardless of map size. This approximation to GP mapping provides high accuracy with limited computational budget for the specialized task of performing fast online map updates and fast online queries of large-scale geospatial maps. It is therefore a suitable approximation for use in real-time applications where such properties are desirable, such as real-time simultaneous localization and mapping (SLAM)  in large, nonlinear geospatial fields. We show on experimental data with magnetic field measurements that our algorithm is faster and equally accurate compared to existing methods, both for recursive magnetic field mapping and for magnetic field SLAM.
\end{abstract}


\section{Introduction}
Navigation in unknown terrains is often performed using on-board sensors measuring the change in position and orientation~\cite{gustafsson_statistical_2013}. However, integrating such measurements causes unbounded error growth in position and orientation estimates over time~\cite{woodman_introduction_2007 ,kok_using_2017-1}. This error growth can be overcome if absolute measurements of the position are included, for example the position relative to a known map. Our focus is on creating such maps using Gaussian process (GP) regression, combining measurements with available prior physical information~\cite{rasmussen_gaussian_2005,wahlstrom_modeling_2015}. Our approach can be used for recursively estimating these maps, as well as for simultaneous localization and mapping (SLAM)~\cite{baileyD:2006,durrantWhyteB:2006}. The latter allows for obtaining position estimates and simultaneously estimating a terrain map to correct for the growing position errors. Examples of terrain maps that can be used to limit this error growth are magnetic field maps~\cite{kok_scalable_2018, robertson_simultaneous_2013, jung_indoor_2015, vallivaara_magnetic_2011}, underwater bathymetry maps~\cite{torroba_online_2023, barkby_bathymetric_2011}, or nonlinear terrain fields~\cite{kjaergaard_terrain_2011, yu_terrain_2018}. 



\begin{figure}
     \centering
     \includegraphics[width=0.25\textwidth,trim={1.1cm 1.2cm 0 0},clip]{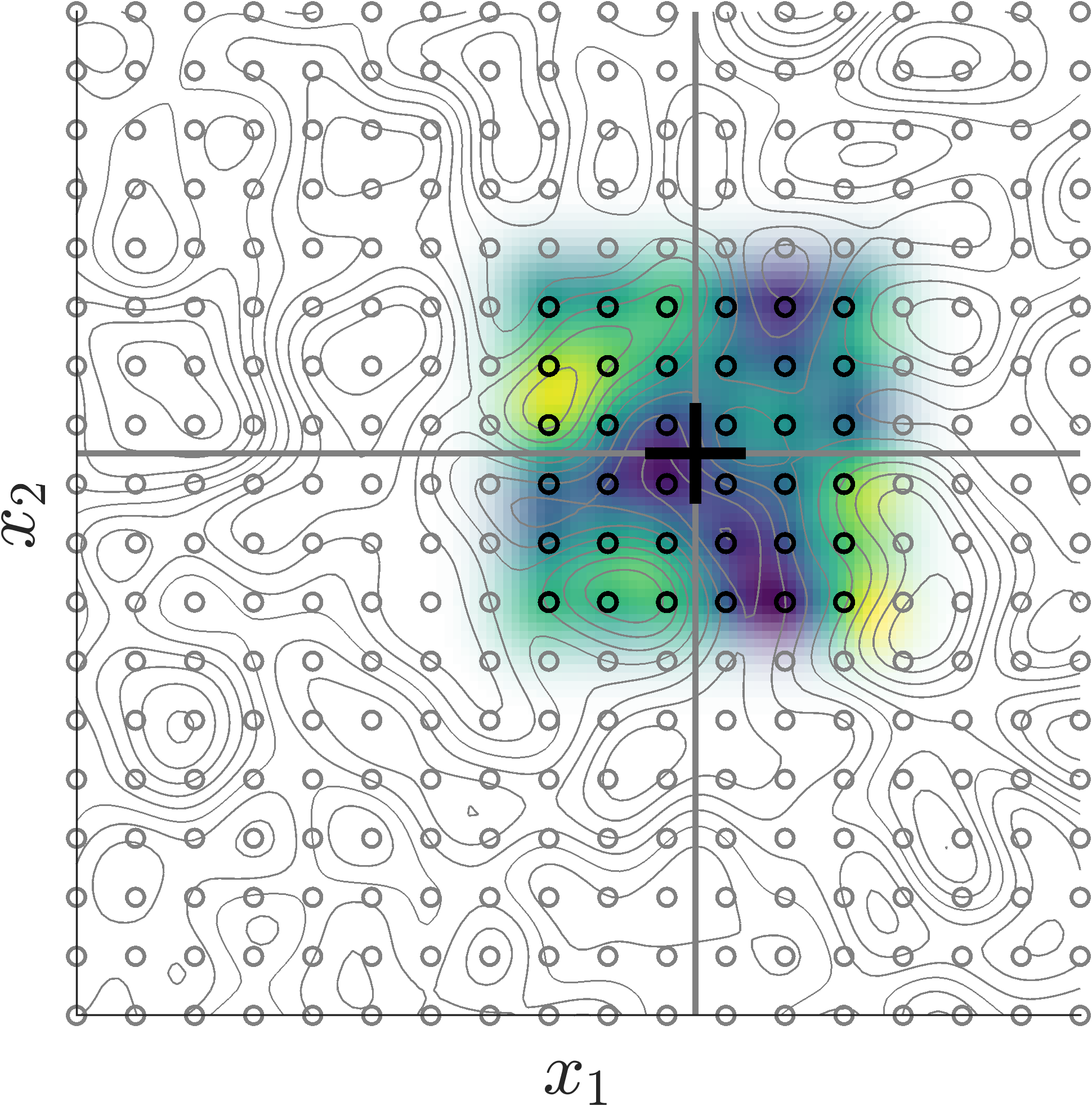}
        \caption{A locally reconstructed approximation (indicated by the color of the heatmap) of a simulated large, nonlinear geospatial field (indicated with gray level curves) based on a local subset (marked with the black circles) of a global grid of basis functions (marked with the gray circles).}
        \label{fig:toy_examples}
\end{figure}

Traditional methods for GP regression face computational and storage challenges, limiting their applicability for real-time applications.
Many existing methods therefore approximate GPs using basis functions~\cite{solin_hilbert_2014, kok_scalable_2018, vallivaara_magnetic_2011, jung_indoor_2014}. Basis function approximations to GPs are for instance used for online creation of magnetic field maps~\cite{kok_scalable_2018}, underwater bathymetry maps~\cite{torroba_online_2023} and ground elevation maps~\cite{viseras_decentralized_2016}.  However, for large-scale fields with small-scale variations a large number of basis functions is needed~\cite{solin_hilbert_2014, ding_multiresolution_2017}. This results in a large computational complexity even for these approximate methods. 
To overcome this limitation, our proposed approximation 1) uses a global grid of finite-support basis functions to represent the map and 2) restricts computations to a local subset of the grid around the region of interest at each timestep. The effect of considering only a local subset of the grid for map approximation is illustrated in Fig.~\ref{fig:toy_examples}. The illustration shows that the map represents the nonlinear field with high confidence in the region of interest, while not spending computational resources on accurately representing the entire field. Our method is inherently recursive, making it suitable both for online creation of GP maps as well as for SLAM. 

The remainder of the paper is organized as follows: Section~\ref{sec:relatedWork} presents related work, while Section~\ref{sec:background} gives an overview of relevant background information. Specifically, it introduces GP regression and basis function approximations to GPs. Section~\ref{sec:method} gives a description of our method for approximating a large GP scale map. It also shows how the map can be incorporated in an extended Kalman filter (EKF) for SLAM, building on~\cite{viset_extended_2022}. Section~\ref{sec:results} presents experimental results comparing our proposed approach to other mapping and SLAM algorithms. Section~\ref{sec:conclusion} gives some concluding remarks and recommendations for future work.

\section{Related work}
\label{sec:relatedWork}
Our interest lies in online construction of a terrain map using a constant stream of incoming measurements. Out-of-the-box GP regression~\cite{rasmussen_gaussian_2005} to construct these maps has a computational complexity that scales with order $\mathcal{O}(N^3)$, where $N$ is the number of measurements. The cost to create the maps hence increases cubically with time. To overcome this limitation, online GP regression typically uses a basis function approximation~\cite{viseras_decentralized_2016, kok_scalable_2018, viset_extended_2022,  jang_multi-robot_2020, viset_magnetic_2021, torroba_online_2023, bijl_online_2015}.\footnote{Note that the widely used inducing point approximations~\cite{quinonero-candela_unifying_2005, hensman_variational_2016} can be seen as a special case of basis function approximations~\cite{viset:2024}.} Using a finite number of $m$ basis functions gives a computational cost at each timestep of $\mathcal{O}(m^2)$. To further mitigate the issue of computational complexity,~\cite{kullberg_online_2021} proposed to use finite-support basis functions in combination with a sparse-weight Kalman filter~\cite{julier:2001}. This approach reduces the computational complexity of including a new measurement to $\mathcal{O}(m)$. These basis functions are placed in a grid-like pattern over the mapped area. The bigger the area, the more basis functions are required to cover it~\cite{solin_hilbert_2014}. This means that as the area increases, the number of basis functions $m$ increases, and the computational complexity $\mathcal{O}(m)$ also increases. However, we still want to be able to create these large maps, because they can aid navigation when moving through the mapped area~\cite{karlsson_bayesian_2006, vasudevan_gaussian_2009, tichavsky_grid-based_2023, gustafsson_particle_2002}. Our approach places a grid of basis functions across the entire mapped area, but only updates values associated with a small local subset of basis functions when receiving new measurements, and only uses a small local subset when making predictions, making it faster than state of the art at both online map creation and online querying of the map. 

Other approaches that overcome the limitations of computational complexity of basis function approximations e.g. split the map into subdomains~\cite{kok_scalable_2018, osman_indoor_2022} or split the measurements into local groups of measurements~\cite{gramacy_local_2015, snelson_local_2007-1, park_efficient_2016, park_domain_2011}. These approaches, however, suffer from boundary-effects. Patched local GPs~\cite{park_efficient_2016} and domain decomposition methods~\cite{park_domain_2011} both remedy this issue by introducing constraints connecting the local domains. However, this remedy to the boundary effect problem requires a number of computations that scales with the number of domains and thus does not truly achieve a prediction time complexity independent of the spatial size of the nonlinear field~\cite{park_efficient_2016}. Other spatially scalable alternatives like KD-tree-based nearest neighbor approaches can include new measurements with a finite computational cost. However, the computational cost for querying the map increases with the number of measurements~\cite{vallivaara_magnetic_2011, vasudevan_gaussian_2009}. In contrast, our approach has a finite computational cost both for including a new measurement, and for querying the map.

Our method is most closely related to a GP approximation called SKI~\cite{pmlr-v37-wilson15}. That method can be used to efficiently include a new measurement in the map using a local subset of basis functions~\cite{yadav_faster_2021}, but has a large computational cost for querying the map. While we use exactly the same strategy as SKI to include new measurements, we propose a new and faster approach to query the map, a procedure which is essential to use the terrain map for navigation purposes. 

\section{Background}\label{sec:background}

\subsection{GP regression}\label{sec:full_GP_regression}

We are interested in estimating a terrain map using GP regression. GP regression allows for estimating a nonlinear function $f:\mathbb{R}^d\rightarrow\mathbb{R}$, distributed according to
\begin{equation}
    f\sim\mathcal{GP}(0,\kappa(\cdot,\cdot)),\label{eq:GP_prior}
\end{equation}
where $\kappa(x,x'):\mathbb{R}^d\times\mathbb{R}^d\rightarrow\mathbb{R}$ is some known kernel function, and $\mathcal{GP}(0,\kappa(\cdot,\cdot))$ denotes the GP prior with a mean of $0$ and covariance defined by the kernel function~\cite{rasmussen_gaussian_2005}. The kernel most commonly used for terrain mapping is the squared exponential kernel defined as
\begin{equation}\label{eq:SE_kernel}
    \kappa_{\text{SE}}(x,x')=\sigma_{\text{SE}}^2 \, \text{exp} \left ( \frac{\|x-x'\|_2^2}{2l_{\text{SE}}^2} \right ),
\end{equation}
where $\left \| \cdot \right \| _2$ is the Euclidean norm, $\sigma_{\text{SE}}$ is a hyperparameter representing the magnitude of the spatial variations and $l_{\text{SE}}$ is a hyperparameter representing the expected lengthscale of the spatial variations. Furthermore, $x$ and $x'$ are two input locations in $
\mathbb{R}^d$~\cite{rasmussen_gaussian_2005}. GP regression uses $N$ noisy measurements of the function $y_{1:N}=\{y_t\}_{t=1}^{N}$ modelled as
\begin{equation}
    y_t=f(x_t)+e_t,\qquad e_t\sim\mathcal{N}(0,\sigma_{\text{y}}^2),
\end{equation}
where $x_{1:N}=\{x_t\}_{t=1}^{N}$ with $x_t\in\mathbb{R}^d$ are known input locations, $e_t$ is measurement noise, $\sigma_{\text{y}}^2$ the noise variance, and $\mathcal{N}(0,\sigma_{\text{y}}^2)$ denotes the normal distribution with mean $0$ and covariance $\sigma_{\text{y}}^2$. The expected value and variance of the function in any arbitrary location $x^{\star}\in\mathbb{R}^{d}$ is then given by
\begin{subequations}
\begin{align}
\begin{split}
    &\expect[f(x^{\star})]\\
    &=K(x^{\star},x_{1:N})\left(  K(x_{1:N},x_{1:N})+\sigma_{\text{y}}^2 I_{N} \right)^{-1}y_{1:N},\label{eq:GP_prediction}\\
\end{split}\\
\begin{split}
    &\text{Var}\left [f(x^{\star})\right]= K(x^{\star},x^{\star}) - \\
    &K(x^{\star},x_{1:N})\left(  K(x_{1:N},x_{1:N})+\sigma_{\text{y}}^2 I_{N} \right)^{-1}K(x_{1:N},x^{\star})\label{eq:GP_var},
\end{split}
\end{align}
\end{subequations}
respectively. Here, the matrix $K(x_{1:N},x_{1:N})$ is constructed by evaluating the kernel along each possible cross-combination of the entries in the vector $x_{1:N}$, such that the entry on the $i$th row and the $j$th column of $K(x_{1:N},x_{1:N})$ is $\kappa(x_i,x_j)$. Similarly, the row vector $K(x^{\star},x_{1:N})$ is defined such that the $j$th column is $\kappa(x^\star,x_j)$. Using the same notation, $K(x^{\star},x^{\star})$ is a single-entry matrix with value $\kappa(x^{\star},x^{\star})$. In this work we approximate the GP posterior from~\eqref{eq:GP_prediction},~\eqref{eq:GP_var}. To distinguish the posterior in~\eqref{eq:GP_prediction},~\eqref{eq:GP_var} from any approximation of it, we refer to it as the full GP posterior. Computing the full GP posterior has a complexity of $\mathcal{O}(N^3)$, as it requires the inversion of a $N\times N$ matrix. 

\subsection{Sparse approximations to GP regression with basis functions}
\label{sec:sparse_approximations}

Sparse approximations to GP regression approximate the function $f$ with a linear combination of $m$ basis functions according to
\begin{equation}\label{eq:general_inducing_functions_model}
    f\approx\Phi^{\top}w, \qquad w\sim\mathcal{N}(0,P),
\end{equation}
where $w=[{w}_{1}, \hdots, w_{m}]^\top$ is a vector of $m$ scalar weights, and $\Phi=[{\phi}_{1}, \hdots, {\phi}_{m}]^\top$ is a vector of $m$ basis functions $\phi_i:\mathbb{R}^{d}\rightarrow\mathbb{R}$~\cite{quinonero-candela_unifying_2005}. The prior covariance on the weights $P\in\mathbb{R}^{m\times m}$ is chosen so that~\eqref{eq:general_inducing_functions_model} approximates~\eqref{eq:GP_prior}~\cite{quinonero-candela_unifying_2005}. 

There exist different methods to approximate predictions using the assumption in~\eqref{eq:general_inducing_functions_model}. A commonly used method is the Deterministic Training Conditional (DTC) approximation~\cite{quinonero-candela_unifying_2005}. The sparse predictions with DTC are given by
\begin{subequations}
\begin{align}
\begin{split}
    \expect\left[{f(x^{\star})}\right]\approx\:&\Phi(x^{\star})^\top\left(\Phi(x_{1:N})\Phi(x_{1:N})^\top+\right.\\
    &\left.\sigma_{\text{y}}^2P^{-1}\right)^{-1}\Phi(x_{1:N})y_{1:N},\label{eq:inducing_point_predictions1}
\end{split}\\
    \begin{split}\text{Var}\left
    [f(x^{\star})\right]\approx\:&\sigma_{\text{y}}^2\Phi(x^{\star})^\top\left(\Phi(x_{1:N})\Phi(x_{1:N})^\top+\right.\\ &\left.\sigma_{\text{y}}^2P^{-1}\right)^{-1}\Phi(x^{\star})\\
    &+K(x^{\star},x^{\star})-\Phi(x^{\star})^\top P\Phi(x^{\star}).\label{eq:inducing_point_predictions2}
    \end{split}
\end{align}
\end{subequations}
We refer to the entry on the $i$th row and the $j$th column of the matrix $\Phi(x_{1:N})$ as $\phi_i(x_j)$, and to the $i$th row of the column vector $\Phi(x^{\star})$ as $\phi_i(x^{\star})$. The expressions in~\eqref{eq:inducing_point_predictions1} and~\eqref{eq:inducing_point_predictions2} require $\mathcal{O}(Nm^2+m^3)$ operations to compute, and $\mathcal{O}(Nm^2)$ storage. This is significantly smaller than the computational cost and storage associated with the full GP predictions~\eqref{eq:GP_prediction}, \eqref{eq:GP_var} when the number of basis functions $m$ is much smaller than the number of data points $N$. The number of basis functions required to accurately approximate the GP scales with the size of the input domain relative to the lengthscale ($l_{\text{SE}}$ in~\eqref{eq:SE_kernel}) of the kernel~\cite{solin_hilbert_2014}. 

\section{Method}\label{sec:method}
In terrain mapping we typically have large areas and small lengthscales. The approach from Section~\ref{sec:sparse_approximations} therefore needs many basis functions resulting in a high computational complexity. To remedy this, we instead propose to use a large grid of finite-support basis functions but only use a local subset of these at each timestep. Section~\ref{sec:choice_basis_functions} introduces the finite-support basis functions. Section~\ref{sec:online_training_IF} describes how we include a new measurement. 

Section~\ref{sec:predictions} describes how we use this trained map to make a prediction. 
 
Section~\ref{sec:SLAM_information_form} presents how our proposed GP map approximation can be integrated into EKF SLAM with GP maps.

\subsection{Choice of basis functions}\label{sec:choice_basis_functions}

We choose a set of truncated basis functions $\{\phi_j\}_{j=1}^{m}$ with centers $u_j\in\mathbb{R}^d$ distributed uniformly on a $d$-dimensional grid as
\begin{equation}\label{eq:finite_support_inducing_functions}
    \phi_j(x)=\left\{\begin{matrix}
\kappa(u_j,x), & \|x-u_j\|_{\infty}\leq r\\ 
0, & \|x-u_j\|_{\infty}>r
\end{matrix}\right.,
\end{equation}
where $\|\cdot \|_{\infty}$ denotes the sup-norm and $r$ the truncation limit. Since the function $\kappa_{\text{SE}}(u_j, x)$ tends to zero as $
\| x-u_j \|_\infty \rightarrow \infty$, this truncation means that values which are close to zero are approximated as exactly zero. This approximation has a low impact on the accuracy when $r$ is chosen large relative to the lengthscale $l_{\text{SE}}$ from~\eqref{eq:SE_kernel}. 

The finite support of our basis functions from~\eqref{eq:finite_support_inducing_functions} ensures that there is always only a finite number of basis functions with overlapping support in any given location. We denote this number of basis functions by $m'$. This results in sparsity in the matrices requiring inversion in~\eqref{eq:inducing_point_predictions1} and~\eqref{eq:inducing_point_predictions2}.

\subsection{Constructing the map}\label{sec:online_training_IF}

The GP approximation in terms of basis functions (see~\eqref{eq:general_inducing_functions_model}) is a parametric model that is linear in the weights $w$. Hence, the posterior of these weights can be found using stochastic least squares. We solve this stochastic least squares problem recursively using an information filter without any dynamics. Specifically, obtaining the posterior on information form corresponds to computing the terms $\Phi(x_{1:N})y_{1:N}$ and $\Phi(x_{1:N})\Phi(x_{1:N})^\top$ in~\eqref{eq:inducing_point_predictions1} and~\eqref{eq:inducing_point_predictions2} recursively~\cite{mutambara_decentralized_1998}. In the remainder of this paper, we let the information vector at time $t$ be defined as $\iota_{1:t}=\Phi(x_{1:t})y_{1:t}$, and the information matrix at time $t$ be defined as $\mathcal{I}_{1:t}=\Phi(x_{1:t})\Phi(x_{1:t})^\top$. The information vector and the information matrix encode the aggregated information from all the available measurements up until and including time $t$. We update the information vector $\iota_{1:t}$ and the information matrix $\mathcal{I}_{1:t}$ as~\cite{mutambara_decentralized_1998}
\begin{subequations}
\begin{align}\label{eq:inducing_functions_information_recursion1}
    \iota_{1:t}=&\iota_{1:t-1}+\Phi(x_t)y_t,\\
    \mathcal{I}_{1:t}=&\mathcal{I}_{1:t-1}+\Phi(x_t)\Phi(x_t)^\top.\label{eq:inducing_functions_information_recursion2}
\end{align}
\end{subequations}
This only requires updating a finite amount of elements $m'\ll m$ in each update step due to the finite support of our basis functions~\eqref{eq:finite_support_inducing_functions}. Specifically, since only $m'$ basis functions have overlapping support in any given location, the terms $\Phi(x_t)y_t$ and $\Phi(x_t)\Phi(x_t)^\top$ only contain $m'$ non-zero elements. Which elements of the information vector and matrices need to be updated can be identified by defining the subset $\mathcal{S}$ of the basis functions from~\eqref{eq:finite_support_inducing_functions} that are non-zero in $x_t$ as
\begin{equation}\label{eq:set_definition}
    \mathcal{S}(x_t,r)=\{j\:|\: \| x_t-u_j\|_{\infty} \leq r\}.
\end{equation}
The updates~\eqref{eq:inducing_functions_information_recursion1},~\eqref{eq:inducing_functions_information_recursion2} only need to be applied to entries $j\in \mathcal{S}(x_t,r)$ for the information vector, and $j,j' \in \mathcal{S}(x_t,r)\times S(x_t,r)$ for the information matrix. The contribution from the term $\phi_j(x_t)y_t$ and $\phi_j(x_t)\phi_{j'}(x_t)$ will inherently be zero for all other entries. 

\subsection{Querying the map - Prediction of terrain values in new locations}\label{sec:predictions}
To use a map constructed using the method from Section~\ref{sec:online_training_IF} for navigation purposes, we need to be able to predict terrain value at new locations. In other words, we need to evaluate the GP posterior mean $E[f(x^*)]$ and variance $\text{Var}[f(x^*)]$ at that point.

In principle, we could straightforwardly use the full information vector ${\iota}_{1:N}$ and information matrix $\mathcal{I}_{1:N}$ in~\eqref{eq:inducing_point_predictions1},~\eqref{eq:inducing_point_predictions2}. However, this requires inversion of the information matrix, an $O(m^3)$ operation. To optimize computation, we instead utilize a smaller, local subset of basis functions, denoted as \(S^*\), that are near the prediction point. This subset \(S^*\) consists of all basis functions within a distance \(r^*\) from the prediction point \(x^*\), as illustrated in Fig.\ 1. Note that the entries of the information vector and matrix can be expressed as
\begin{align}
        \mathcal{I}^{i,j}_{1:N} = \sum_{t=1}^{N} \phi_i(x_t) \phi_j(x_t), \quad      \iota^i_{1:N} = \sum_{t=1}^{N} \phi_i(x_t) y_t,  
\end{align}
where $i\in{1,\hdots,m}$ and $j\in{1, \hdots, m}$ are indices corresponding to each entry of the information matrix. Hence, using this property, we define the local information vector ${\iota}^*_{1:N}$ and matrix $\mathcal{I}^*_{1:N}$ as
\begin{subequations}
    \begin{align}
        {\iota}^*_{1:N} &= \Phi_{S^*}(\mathbf{x}_{1:N}) \mathbf{y}_{1:N}, \\
        \mathcal{I}^*_{1:N} &= \Phi_{S^*}(\mathbf{x}_{1:N}) \Phi_{S^*}(\mathbf{x}_{1:N})^\top,
    \end{align}
\end{subequations}
where \(\Phi_{S^*}(\mathbf{x})\) represent the evaluations of the basis functions in the local subset \(S^*\) at any location \(\mathbf{x}\).
We now use ${\iota}^*_{1:N}$, $\mathcal{I}^*_{1:N}$ to approximate the predicted mean and variance as
\begin{subequations}
\begin{align}
    E[f(x^*)] \approx & \Phi_{S^*}(x^*)^\top (\mathcal{I}^*_{1:N} + \sigma_\text{y}^2 {P}^*)^{-1} {\iota}_{S^*}, \\
    \text{Var}[f(x^*)] \approx & \sigma_\text{y}^2 \Phi_{S^*}(x^*)^\top (\mathcal{I}^*_{1:N} + \sigma_\text{y}^2 {P}^*)^{-1} \Phi_{S^*}(x^*) \nonumber \\
    &  + K(x^*, x^*) - \Phi_{S^*}(x^*)^\top {P}^* \Phi_{S^*}(x^*).
\end{align}
\label{eq:predictionLocalFinite}%
\end{subequations}%
Here, \({P}^*\) is the prior covariance of the basis function weights, set to recover the prior accurately at the centers of the selected basis functions. The detailed derivation of \({P}^*\) is provided in Appendix A. Note that the maximum size of the local subset \(S^*\) is $
m'' \leq (\tfrac{2r^*}{l_u} + 1)^d$, where \(d\) is the dimension of the input vector. The computational complexity for computing~\eqref{eq:predictionLocalFinite} is therefore \(O(m''^3)\).

\subsection{Integration of mapping algorithm into an EKF for magnetic field SLAM}\label{sec:SLAM_information_form}
Our approach allows for constructing a map (see Section~\ref{sec:online_training_IF}) and for querying the map to use it for navigation (see Section~\ref{sec:predictions}). In this section we combine both, and show that our approach can also be used for SLAM. We illustrate this for magnetic field SLAM using an extended Kalman filter (EKF), which we refer to as EKF Mag-SLAM. 
We specifically modify the algorithm presented in~\cite{viset_extended_2022}, but use our proposed map approximation instead of the Hilbert space (HS) basis functions used in~\cite{viset_extended_2022}. 

Our task for SLAM is to estimate the joint posterior distribution:
\[
p(p_t, q_t, w_t \mid y_{1:t}),
\]
where $p_t$ is a three-dimensional position vector, $q_t$ is a 4-dimensional unit quaternion representing the orientation, and $w_t$ is an $m$-dimensional vector of weights associated with finite-support basis functions \(\phi_i(p)\) representing the magnetic field. This posterior is approximated in the EKF framework using a Gaussian distribution. To deal with the fact that orientations are non-Euclidean, we implement an error-state EKF. This implies that the EKF not only consists of a dynamic update and a measurement update, but also a re-linearization step. In line with Sections~\ref{sec:online_training_IF},~\ref{sec:predictions}, we implement this EKF on information form. This means that instead of keeping track of the state estimates $\hat{p}_{t|t}$, $\hat{q}_{t|t}$, $\hat{m}_{t|t}$ and their corresponding covariances, we keep track of the corresponding information matrix $\mathcal{I}^{\text{EKF}}_{t|t}$ and information vector $\iota^{\text{EKF}}_{t|t}$.

\subsubsection{State-space model}
Let us assume that information about the change in position and orientation between two time steps is available, e.g.\ from wheel encoders or inertial sensors. Furthermore, let us assume that the magnetic field map is constant over time. The dynamic model of our states is then 
\begin{subequations}
\begin{align}
    p_{t+1}&=p_{t}+\Delta p_t + e_{\text{p},t},& e_{\text{p},t}\sim\mathcal{N}(0,\sigma_{\text{p}}^2 \mathcal{I}_{3}),\label{eq:odom_pos}\\
    q_{t+1}&=q_t\odot\Delta q_t\odot \exp_{\text{q}}(e_{\text{q},t}),& e_{\text{q},t}\sim\mathcal{N}(0,\sigma_{\text{q}}^2 \mathcal{I}_3),\label{eq:odom_orient}\\
    w_{t+1} &= w_t.
\end{align}
\label{eq:dynModelMagSLAM}%
\end{subequations}%
Here, $\odot$ is the quaternion product and $\exp_{\text{q}}$ is the operator that maps an axis-angle orientation deviation to a quaternion (see \cite{kok_using_2017-1} for details on quaternion algebra). 

Our map of the magnetic field, expressed in a world-fixed frame, assumes that the field is curl-free. It can therefore be expressed as the gradient of a scalar potential $\varphi(p_{t})$ with respect to the position $p_{t}$~\cite{wahlstrom_modeling_2013}. We model the scalar potential $\varphi:\mathbb{R}^3\rightarrow \mathbb{R}$ as a GP with prior
\begin{equation}
    \varphi\sim\mathcal{N}(0,\kappa_{\text{SE}}(\cdot,\cdot)+\kappa_{\text{lin}}(\cdot,\cdot))\label{eq:GP_prior_mag},
\end{equation}
where $\kappa_{\text{lin}}(\cdot,\cdot))$ is a linear kernel~\cite{rasmussen_gaussian_2005} used to model the static Earth's magnetic field. The squared exponential kernel is used to model the local magnetic field anomalies. 

We assume that measurements $y_t \in \mathbb{R}^3$ of the magnetic field expressed in a sensor-fixed frame are available. Approximating the scalar potential field in terms of the finite-support basis functions from Section~\ref{sec:choice_basis_functions} results in the measurement model
\begin{equation}\label{eq:meas_model_mag_weightspace}  y_{t}=R_t\nabla_{\text{p}}\Phi(p_{t})^\top w_t+e_{\text{m},t},\qquad e_{\text{m},t}\sim\mathcal{N}(0,\sigma_{\text{m}}^2\mathcal{I}_3).
\end{equation}
Here, $e_{\text{m},t}$ denotes the zero-mean measurement noise with covariance $\sigma_{\text{m}}^2\mathcal{I}_3$, and $R_t$ the rotation from the world-fixed to the sensor-fixed frame. 

In line with existing SLAM literature, we use choose the prior on the position and orientation to be zero-mean with a very small covariance. The GP prior results in a prior mean of zero for the states $w_t$ with a covariance based on the GP prior~\cite{viset_extended_2022}. 

\begin{figure}
     \centering
     \subfloat[1D grid]{\includegraphics[width=1in]{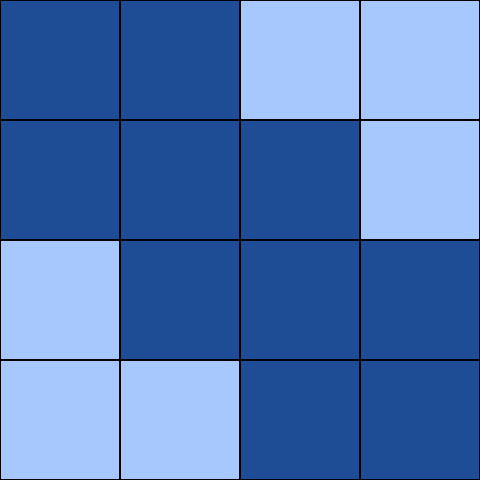}%
    \label{fig_sparsity1D}}
    \hfil
     \subfloat[2D grid]{\includegraphics[width=1in]{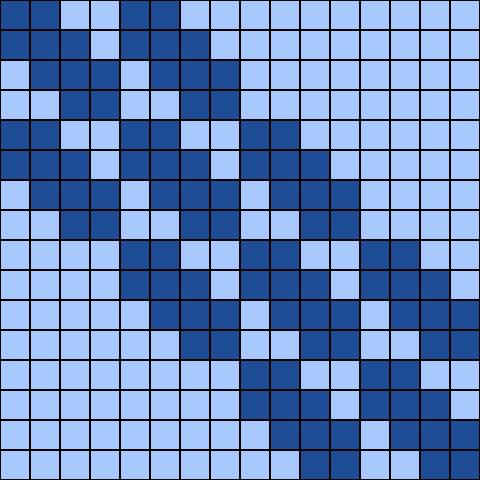}%
    \label{fig_sparsity2D}}
    \hfil
         \subfloat[3D grid]{\includegraphics[width=1in]{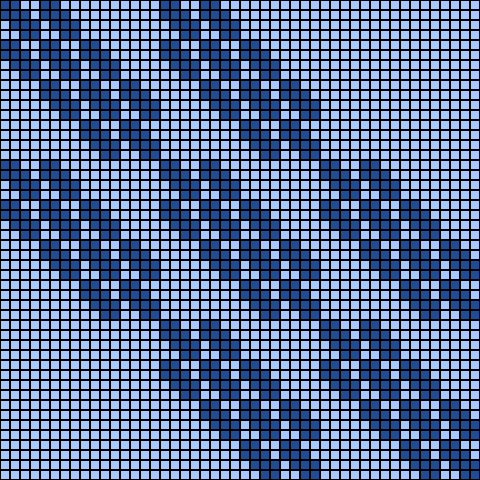}%
    \label{fig_sparsity3D}}
        \caption{Sparsity patterns illustrating which entries $i,j$ in the information matrix correspond to pairs of basis function locations $x_i,x_j$ that are closer than $2r^{\star}$ according to the infinity norm (dark blue) and which are not (light blue). The patterns arise from the ordering of the indexes of the basis functions, relative to their locations along each of the three dimensions.\label{fig:sparsity_patterns_increasing_dimension}}  
\end{figure}

\begin{figure}
    \centering
    \begin{equation*}
\includegraphics[valign=c,width=0.13\textwidth]{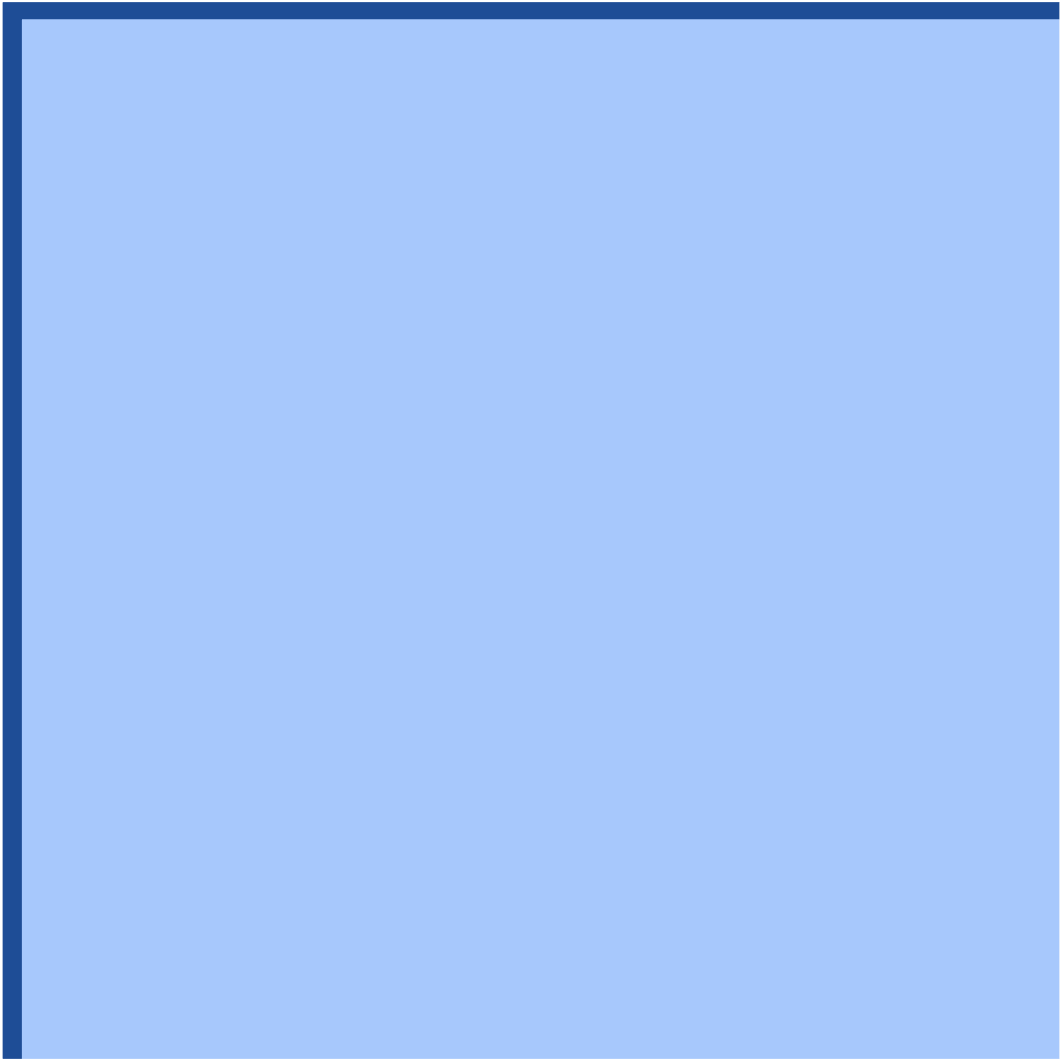} \: \bigcup  \: \includegraphics[valign=c,width=0.13\textwidth]{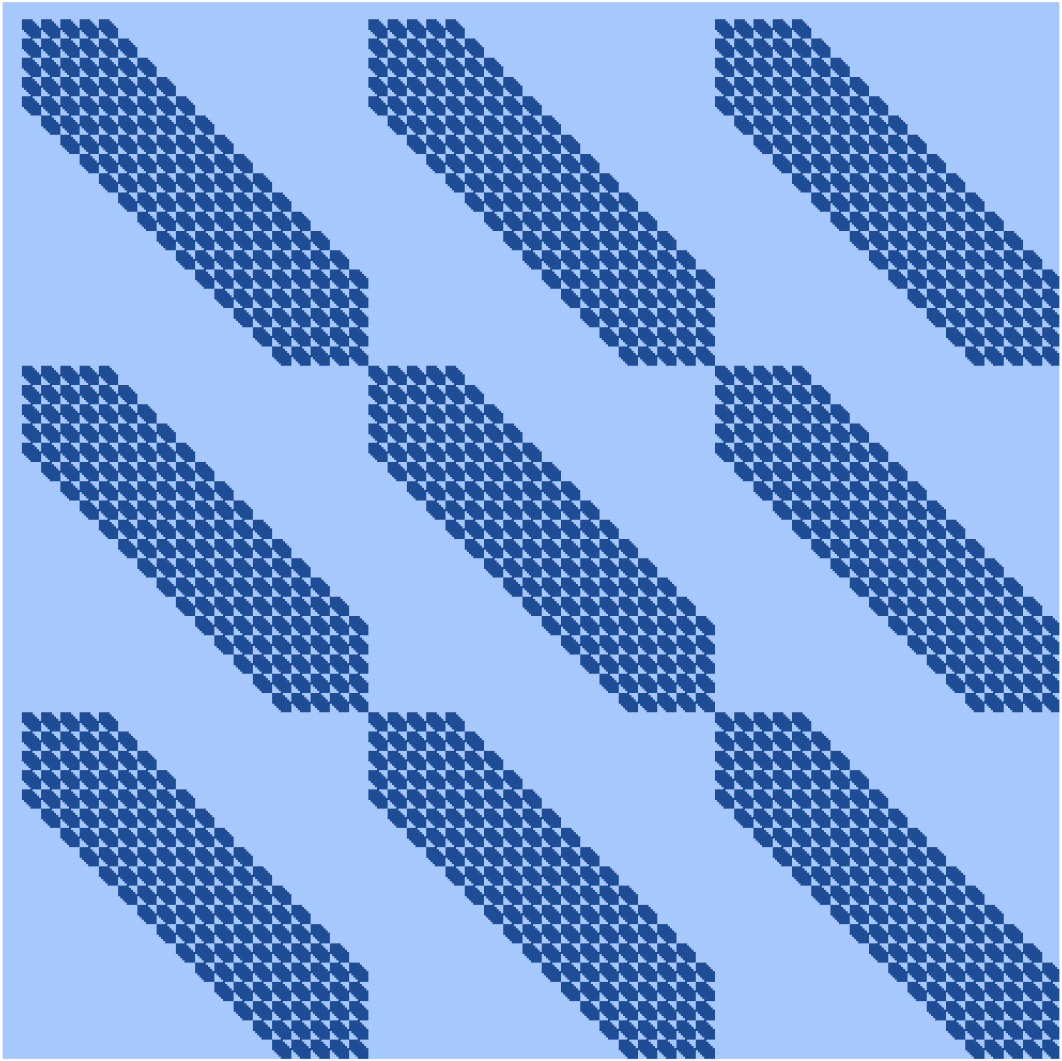} = \includegraphics[valign=c,width=0.13\textwidth]{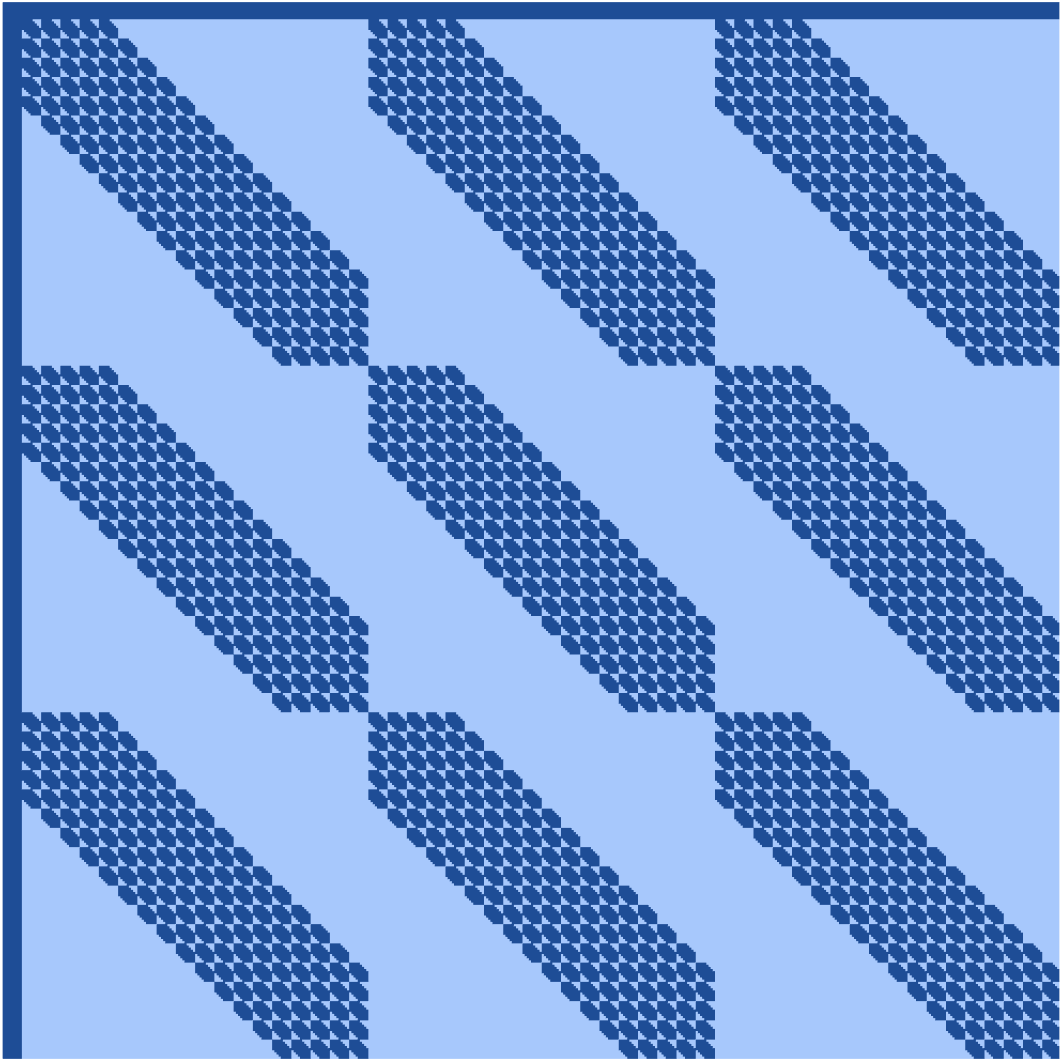} \:.
\end{equation*}
    \caption{Sparsity pattern illustration of the information matrix for the full state consisting of the position, orientation and magnetic field. The dark blue entries indicate entries which are necessary to compute with our assumptions, and the light blue entries indicate values that are not necessary to compute. The first sparsity pattern indicate the values we need associated with the position and orientation (the first 6 states in the full state-space). The second sparsity pattern has a dark blue color in the entries in the set $S_{\forall \star}$ of all entries that can possibly be necessary to make a map prediction in any location. The last sparsity pattern is the union of these two sets.}
    \label{fig:sparsity_illustration}
\end{figure}

\subsubsection{Dynamic update}

The dynamic update on information form~\cite{walter_exactly_2007} is given by
\begin{equation}\label{eq:dynamic_update_information_form}
    \mathcal{I}^{\text{EKF}}_{t+1|t}=(\mathcal{I}^{\text{EKF}}_{t|t}+Q^{-1})^{-1},
\end{equation}
where $Q$ is the process noise covariance for the model~\eqref{eq:dynModelMagSLAM}. Since we assume the map to be static, the matrix $Q$ can be factorized according to
\begin{equation}
    \begin{bmatrix}
\tilde{Q} & 0 \\ 
0 & 0
\end{bmatrix}=V^\top QV, \qquad 
V=\begin{bmatrix}
I & 0
\end{bmatrix},
\end{equation}
where $\tilde{Q}$ is a $6\times6$ matrix representing the process noise on the position and orientation.\footnote{Note that the error state representing the orientation is three-dimensional.}
After applying the matrix inversion lemma to~\eqref{eq:dynamic_update_information_form}, it reduces to
\begin{equation}   \mathcal{I}^{\text{EKF}}_{t+1|t}=\mathcal{I}^{\text{EKF}}_{t|t}-\mathcal{I}^{\text{EKF}}_{t|t}V^\top(V\mathcal{I}^{\text{EKF}}_{t|t}V^\top+\tilde{Q}^{-1})^{-1}VI^{\text{EKF}}_{t|t}.
\end{equation}

\subsubsection{Measurement update}
The Kalman filter measurement update EKF Mag-SLAM is given by~\cite{walter_exactly_2007}
\begin{subequations}
\begin{align}
\iota_{t|t}^\text{EKF}&=\tfrac{1}{\sigma_\text{m}^2}H_ty_t, \label{eq:magSLAMmeasUpdateState} \\
\mathcal{I}^{\text{EKF}}_{t|t}&=\mathcal{I}^{\text{EKF}}_{t|t-1}+\tfrac{1}{\sigma_\text{m}^2}H_tH_t^\top, \label{eq:magSLAMmeasUpdateInf}
\end{align}%
\end{subequations}%
where the $H_t$ is the Jacobian of the measurement model~\eqref{eq:meas_model_mag_weightspace}. Note the absence of a summation in~\eqref{eq:magSLAMmeasUpdateState} due to the error-state implementation of the EKF. The update of the information matrix in~\eqref{eq:magSLAMmeasUpdateInf} is inherently sparse due to our use of finite-support basis functions. This is similar to the sparsity in~\eqref{eq:inducing_functions_information_recursion2},  and has a complexity of $O(m'')$. 

\subsubsection{Re-linearization}
The re-linearisation of the estimated position, orientation and magnetic field of our error-state EKF requires evaluation of the error-state. For this, an inversion of the information matrix is essential. We instead perform an approximate re-linearization by only using the basis functions that are close to the prediction point (which in this case is the estimated location). This allows us to execute the re-linearisation at a computational cost of $\mathcal{O}(m''^3)$ as
\begin{subequations}\label{eq:relinearisation_EKF_SLAM}
\begin{align}
    \begin{bmatrix}
\delta_t ^\top & \eta_t^\top & \nu_{\star,t}^\top
\end{bmatrix}^{\top}&=(I_{t,\star}^{\text{EKF}})^{-1}\iota_{t,\star}^{\text{EKF}},\\
    \hat{p}_{t|t}&=\hat{p}_{t|t-1}+\delta_t,\\
    \hat{q}_{t|t}&=\hat{q}_{t|t-1}\odot\exp_{\text{q}}(\eta_t),\\
    \hat{w}_{\star,t|t}&=\hat{w}_{\star,t|t-1}+\nu_{\star,t}.
\end{align}
\end{subequations}
Here, $\delta_t$, $\eta$ are the error states for the position and orientation, respectively. Furthermore, $\nu_{\star,t}$ is the error state corresponding to the weights of the basis functions in the subset $S_{\star}$, see also~\eqref{eq:set_definition} and Section~\ref{sec:predictions}. In other words, we only correct the linearisation point of a local subset of the magnetic field map.

\subsubsection{Note on computational complexity}
As shown above, the computational complexity of the measurement update of our EKF Mag-SLAM algorithm is $\mathcal{O}(m'')$ and of the relinearization is $\mathcal{O}(m''^3)$. In principle, the dynamic update would have a computational complexity of $\mathcal{O}(m^2)$. However, there are large portions of the matrix $I^\text{EKF}_t$ that never need to be explicitly computed. The only entries of $I_{t}^\text{EKF}$ that need to be computed, are entries that affect the output of the relinearization, the measurement update, or the dynamic update.

Firstly, we will describe the set of entries in $I_{t,\star}^\text{EKF}$ that affect the output of the relinearization and the measurement update. There are many basis function pairs $i$, $j$ that are so far away from each other that they are never used at the same time in the relinearization step, nor in the measurement update. We denote the set of all pairs of basis functions $i$, $j$ that are \emph{close enough} to each other that we \emph{do need} to compute the corresponding entries in $I^\text{EKF}_{t,\star}$ by $S_{\forall \star}$. We can formally define this set of nearby index pairs as $S_{\forall\star}=\{ i,j|\|p_{i,t}-p_{j,t}\|_{\infty}\leq 2r^{\star}\}$. This definition includes all index pairs $i,j$ where the corresponding basis functions are closer to each other according to the infinity norm than $2r^{\star}$. The set $S_{\forall \star}$ contains $\mathcal{O}(mm''^2)$ elements. Assuming $m>>m''$, we simplify this notation, and say that $S_{\forall \star}$ contains $\mathcal{O}(m)$ elements. Fig.~\ref{fig:sparsity_patterns_increasing_dimension} illustrates examples of where the entries $i,j$ in $S_{\forall \star}$ are located in an information matrix corresponding to a one-dimensional, a two-dimensional, and a three-dimensional grid of equispaced basis functions. In the one-dimensional grid, four basis functions are located along the x-axis at $[-1.5, -0.5, 0.5, 1.5]$, and the indices are ordered correspondingly in Fig.~\ref{fig:sparsity_patterns_increasing_dimension}. In the two-dimensional grid, 16 basis functions are located along the x and y-axis at
\begin{equation}
    \begin{bmatrix}
-1.5 & -1.5 & -1.5 & \hdots & 1.5 & 1.5 \\ 
-1.5 & -0.5 & 0.5 & \hdots & 0.5 & 1.5 
\end{bmatrix},
\end{equation}
and also indexed chronologically. The sparsity pattern shows a fractal-like structure that is explained by the fact that we are displaying the pattern for a 3D grid of basis functions, and the nature of the ordering of the indexes of these basis functions. In all cases, only $\mathcal{O}(m)$ elements are part of the set $S_{\forall \star}$, meaning that only $\mathcal{O}(m)$ elements need to be computed in the information matrix to carry out the relinearization and the measurement update.

Secondly, we describe the set of entries in $I_{t,\star}^\text{EKF}$ that affect the output of the dynamic update. The dynamic update would, in general, require knowledge of the entire information matrix. In our case, the only entries of the information matrix that are used to compute the term $\mathcal{I}^{\text{EKF}}V^\top(V\mathcal{I}^{\text{EKF}}V^\top+\tilde{Q}^{-1})^{-1}V\mathcal{I}^{\text{EKF}}$ are the entries in the first 6 rows and the first 6 columns of the information matrix. The only entries of the information matrix we need to keep track to ensure the output of the dynamic update is correct of are therefore the union of the first 6 columns and the first 6 rows and the values in $\mathcal{S}_{\forall \star}$. Fig.~\ref{fig:sparsity_illustration} shows an illustration of the set of indices necessary to perform the dynamic update (the leftmost matrix), the indices necessary to perform the measurement update and the relinearisation (the middle matrix) and the union of these two sets (the rightmost matrix). The union of the sets of indices are all possible indices necessary to get a correct output from the measurement update, the relinearisation and the dynamic update. In other words, the union of these sets of indices are the only entries of the information matrix $I_{t}^\text{EKF}$ we need to compute to  for our EKF Mag-SLAM algorithm. 

In conclusion, an overview of the computational complexities of EKF Mag-SLAM with our mapping technique compared to EKF Mag-SLAM with Hilbert Space basis fucntions from~\cite{viset_extended_2022} is given in Table~\ref{tab:complexitiesEKF}. When $m'$ and $m''$ are chosen to be considerably smaller than $m$, our proposed algorithm will be faster than the approach from~\cite{viset_extended_2022}. This is a reasonable choice when the terrain is large relative to the spatial variations we wish to map.

\begin{table}[h]
\caption{Comparison of computational complexities of EKF Mag-SLAM.}
\centering
\sisetup{ table-align-uncertainty=true,
retain-zero-uncertainty=true,
separate-uncertainty=true,
scientific-notation = false}
\setlength\tabcolsep{5pt}
\robustify\bfseries
\begin{tabular}{
c
S[detect-weight,table-format=1.3(4)]
S[detect-weight,table-format=1.3(4)]
S[detect-weight,table-format=1.3(4)]
S[detect-weight,table-format=1.3(4)]
S[detect-weight,table-format=1.3(4)]
S[detect-weight,table-format=1.3(4)]
S[detect-weight,table-format=1.3(4)]
S[detect-weight,table-format=1.3(4)]
S[detect-weight,table-format=1.3(4)]
S[detect-weight,table-format=1.3(4)]
}
{} &  {EKF} & {EKF} \\
{} &  {Mag-SLAM with} & {Mag-SLAM with} \\
{Step} &  {our method} & {Hilbert Space functions}  \\
\hline
Meas update & {$\mathcal{O}(m'^2)$}  & {$\mathcal{O}(m^2)$}  \\
Prediction & {$\mathcal{O}(m''^3)$} & {$\mathcal{O}(m)$} \\
Dyn update & {$\mathcal{O}(m)$} & {$\mathcal{O}(m^2)$} \\
\hline
\end{tabular}
\label{tab:complexitiesEKF}
\end{table}

\begin{table*}[!t]
\renewcommand{\arraystretch}{1.2}
\caption{Standardised mean absolute errors (SMAEs), combined time to include a new measurement and make a new prediction, mean standardized log-likelihood (MSLL) scores for contructing the sound map. The results that attain the lowest run-time while also attaining the lowest SMAE are highlighted. SMAEs and MSLL scores are evaluated on all standardized test points inside the considered domain, and the average time plus minus one standard deviation is calculated based on 100 repetitions. \label{tab:SMAEaudio}}
\centering
\begin{tabular}{|c|c|c|c|c|c|c|}
\cline{2-7}
\multicolumn{1}{c|}{} & \multicolumn{3}{c|}{10\% of domain, $m=800$} & \multicolumn{3}{c|}{100\% of domain, $m=8000$}\\
\cline{2-7}
\multicolumn{1}{c|}{} & \multicolumn{1}{c|}{SMAE} & Time[s] & MSLL & SMAE & Time[s] & MSLL\\
\hline
$r=6l_{\text{SE}}$ & $0.42$ & $6.2\cdot 10^{-5}\pm 2.1\cdot 10^{-5}$ & $37.5$ & $0.34$ & {$7.3\cdot 10^{-5}\pm2.2\cdot10^{-5}$} & {$48.0$}\\
\hline
$r=12l_{\text{SE}}$ & $\mathbf{0.22}$ & {$\mathbf{9.4\cdot 10^{-5}\pm 2.1\cdot 10^{-5}}$} & $4.18$ & $\mathbf{0.20}$ & $\mathbf{1.1\cdot 10^{-4}\pm1.6\cdot10^{-5}}$ & $12.2$\\
\hline
$r=18l_{\text{SE}}$ & $0.22$ & $1.3\cdot 10^{-4}\pm 2.7\cdot 10^{-5}$ & $3.94$ & $0.20$ & $1.6\cdot 10^{-4}\pm2.1\cdot10^{-5}$ & $12.2$\\
\hline
SKI & $0.22$ & $8.0\cdot10^{-4} \pm 1.2\cdot 10^{-4}$ & $5.74$ & $0.20$ & $1.6\cdot 10^{-3}\pm1.7\cdot10^{-4}$ & $18.2$\\
\hline
Inducing inputs & $0.22$ & $1.0\cdot 10^{-2}\pm 7.4\cdot 10^{-4}$ & $3.81$ & $0.20$ & $2.1\cdot 10^{-1}\pm1.6\cdot10^{-3}$ & $12.2$\\
\hline
Hilbert space & $0.22$ & $9.5\cdot 10^{-3}\pm 5.1\cdot 10^{-4}$ & $3.80$ & $0.20$ & $2.2\cdot 10^{-1}\pm4.6\cdot10^{-3}$ & $12.2$\\
\hline
\end{tabular}
\end{table*}

\section{Results}\label{sec:results}

In this section, we first compare the performance of our method with existing approaches on two low-dimensional benchmark data sets. We also study how long it takes for our method to make online predictions using a short length scale and millions of basis functions on a bathymetry dataset that is too large for existing approaches given our hardware constraints. 

As the data sets considered in Sections~\ref{sec:precip_results} and~\ref{sec:bathymetry_data} are geospatial data with a non-zero mean value, the average of the output is subtracted before training. This average is subsequently added to each prediction. All computation times reported are measured while running on a Dell XPS 15 9560 laptop, with $16$ GB RAM and an Intel Core i7-7700HQ CPU running at 2.80 GHz. In all experiments we set $r=2r^{\star}$, as picking $r>=2r^{\star}$ gives that the expression for $P_{\star}^{-1}$ reduces to $P_{\star}^{-1}=K(u_{S^\star},u_{S^\star})$, as derived in Appendix~\ref{sec:P_star_derivation}. This means that we have a closed-form expression for the inverse of the prior covariance which does not rely upon computing any numerical inverses, further reducing the necessary computational efforts to make each prediction. All code and files required to reproduce the results can be found on~\href{https://github.com/fridaviset/FastGPMapping}{https://github.com/fridaviset/FastGPMapping}.

\begin{figure}
    \centering
    \includegraphics[width=0.45\textwidth]{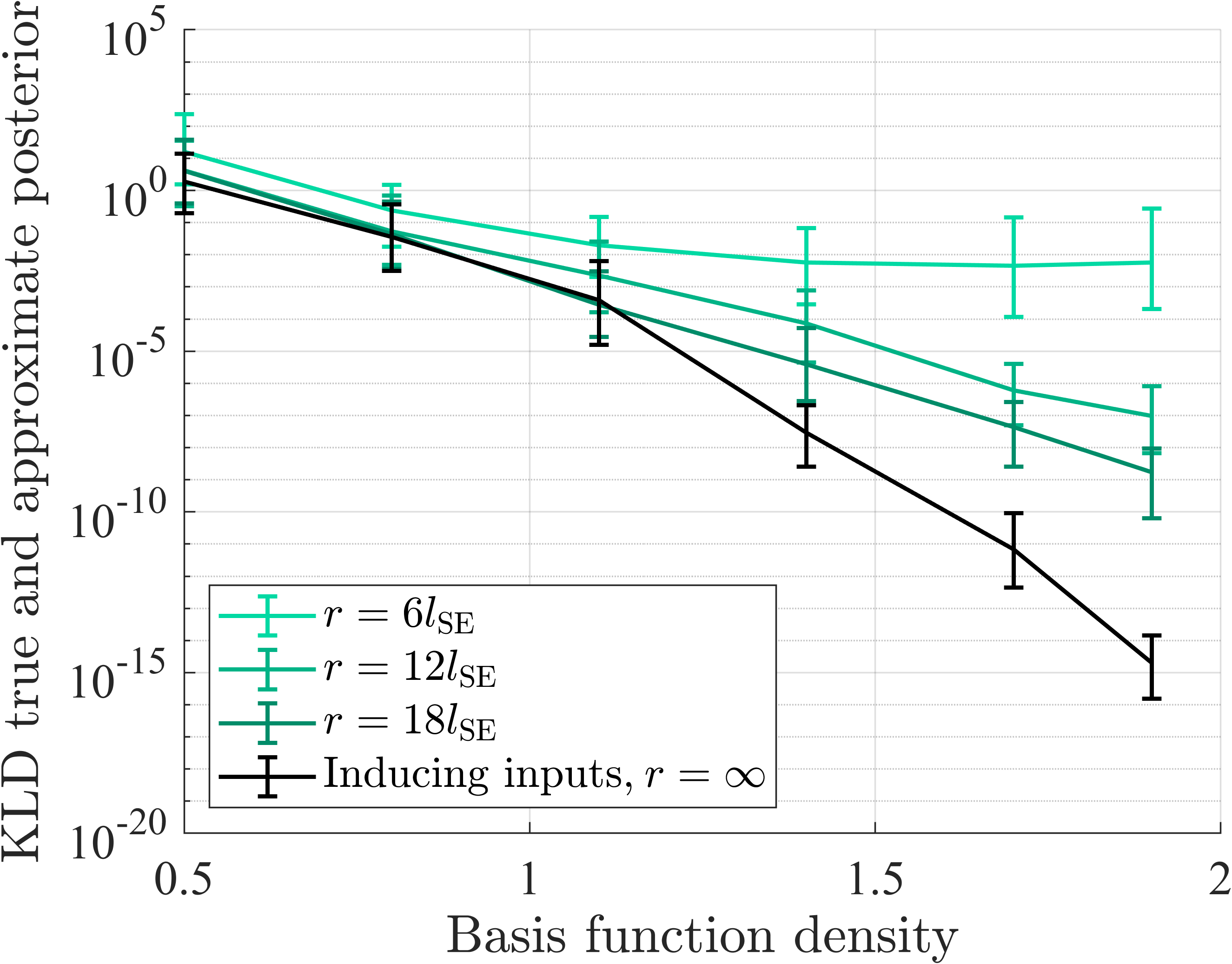}
    \caption{KL divergence (KLD) between the full GP posterior, and approximations with various local domain sizes $r$, trained on the audio dataset. The error bars indicate the average deviation above and below the mean, respectively, after 100 repeated experiments with 100 randomly sampled measurements from the training set.}
    \label{fig:audio_convergence}
\end{figure}

\begin{figure}
         \centering
         \includegraphics[width=0.48\textwidth]{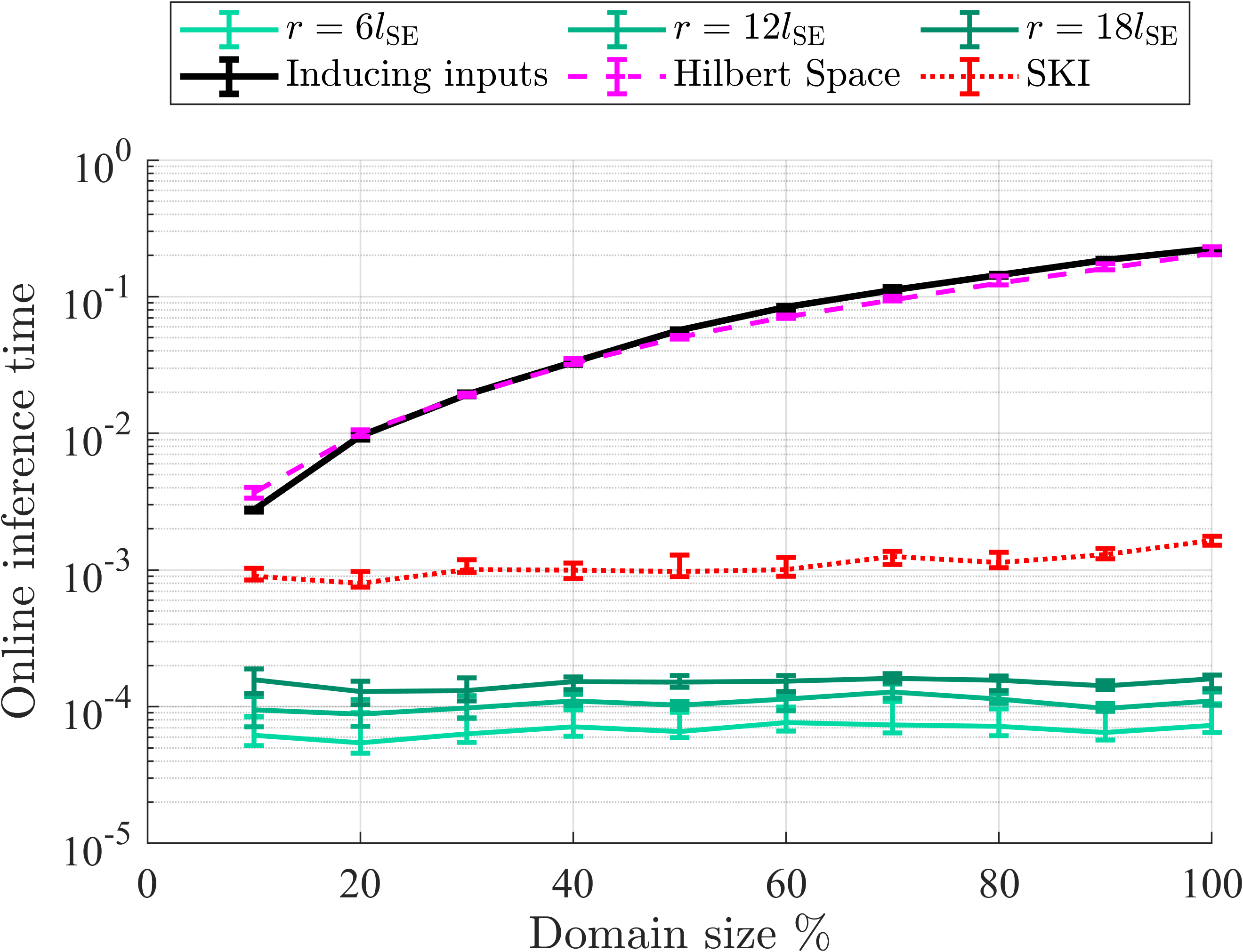}
         \caption{Online inference time of the sound map for a growing domain size. We compare our proposed method (with various local domain sizes $r$) to the inducing input approximation using inducing inputs on a grid, the Hilbert space basis function approximation, and SKI. All methods were run using the same amount of basis functions. The error bars indicate the average deviation above and below the mean after 100 repeated experiments, respectively.}
         \label{fig:times_audio}
\end{figure}

\subsection{One-dimensional sound map}\label{sec:audio_results}

In this section we will construct a one-dimensional GP map of the amplitude of a sound wave using the data from~\cite{turner_statistical_2010}. We treat time as the spatial axis of this map. The same data and approach was used in~\cite{pmlr-v37-wilson15} and~\cite{yadav_faster_2021} to demonstrate the ability of their approximations to learn a GP model from a large dataset with a large input domain relative to the size of the spatial variations in the field. The dataset contains a training set of $59\:309$ measurements of sound amplitude collected at a known input time and a fixed test set of $691$ points. We compare the accuracy and required computation time of our method with existing work. To this end, similar to~\cite{yadav_faster_2021} we use a squared exponential kernel~\eqref{eq:SE_kernel} with hyperparameters $\sigma_{\text{SE}}=0.009$, $l_{\text{SE}}=10.895$, $\sigma_{\text{y}}=0.002$. 

In Fig.~\ref{fig:audio_convergence} we investigate how large the size of the local domain $r$ (see Section~\ref{sec:choice_basis_functions}) has to be for our approach to accurately approximate the posterior. We assess our results in terms of the KL divergence between the approximate posterior and the full GP posterior. The KL divergence is a measure that compares how similar distributions are. It can attain values between 0 and $\infty$, where a lower value means that the distributions are more similar. First of all, it can be seen that a higher basis functions density (measured in the number of basis functions per lengthscale) results in a smaller KL divergence, i.e.\ a better approximation, of the inducing input approximation. Our method can be seen to approach the inducing input approximation for larger $r$. 

To measure how long it takes for our approach to perform online mapping, we measure the time it takes to include one additional measurement and perform one prediction step. In Fig.~\ref{fig:times_audio}, we compare our proposed method to the inducing input solution implemented with an online Kalman filter~\cite{bijl_online_2015}, an online Kalman filter implementation with Hilbert space basis functions~\cite{solin_hilbert_2014}, and an online implementation of SKI~\cite{yadav_faster_2021}, for increasing domain sizes. As the domain size increases, we keep the basis function density constant to retain the same approximation accuracy. The amount of basis functions $m$ is therefore increasing linearly.  As the online inference time of SKI is affected by how many iterations are used in the conjugate gradient solver to improve the approximation accuracy, we measure the run-time using just one iteration in the solver. To compare the prediction accuracy of SKI to our approach in Table~\ref{tab:SMAEaudio}, however, we use the exact solution that the solver is approximating. We can therefore conclude that an online evaluation of our algorithm is faster than competing approaches while being able to recover the same or better SMAE (standardized mean average error) and MSLL (mean squared log loss) scores (see Table~\ref{tab:SMAEaudio}). The online computational complexity of both Hilbert space basis functions and inducing inputs increases quadratically as the domain size increases. The computational complexity of our approach is bounded by $O(m'^3)$ independent of the increase in the domain size, which is why our online computational complexity remains lower than $10^{-3}$ seconds independent of the growth of the domain size. 

\begin{table*}[!t]
\renewcommand{\arraystretch}{1.2}
\caption{SMSE accuracies of daily precipitation level predictions. The predictions are obtained with a local domain with size $r^{\star}=3l_{\text{SE}}$, which corresponds to using at most 144 basis functions in each local prediction for the highest $m=200K$. \label{tab:SMSEs}}
\centering
\begin{tabular}{|c|c|c|c|c|c|c|c|}
\cline{3-8}
\multicolumn{2}{l|}{}  & \multicolumn{2}{c|}{Inducing inputs} & \multicolumn{4}{c|}{Local information filter} \\ 
\hline
{$N$} & full GP & m=10K & m=20K & m=10 K & m=20K & m=100K & m=200K \\
\hline
$10$ $000$ & $0.823$ & $0.957$ & $0.905$ & $0.957$ & $0.906$ & $0.824$ & $0.823$ \\
\hline
$20$ $000$ & $0.766$ & $0.946$ & $0.861$ & $0.947$ & $0.862$ & $0.770$ & $0.766$ \\
\hline
$100$ $000$ & N/A & $0.907$ & $0.782$ & $0.907$ & $0.786$ & $0.561$ & $0.545$ \\
\hline
$528$ $474$ & N/A & $0.894$ & $0.746$ & $0.895$ & $0.751$ & $0.468$ & $0.435$\\
\hline
\end{tabular}
\end{table*}

\begin{figure*}
     \centering
     \subfloat[$m=2.33$M, $N=373$K.]{\includegraphics[width=2in]{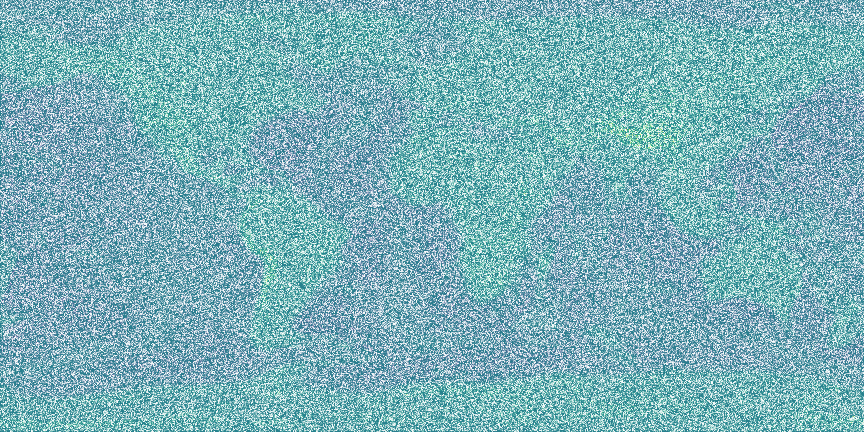}%
    \label{fig:bathymetry1}}
    \hfil
         \subfloat[$m=286$K, $N=37.3$M.]{\includegraphics[width=2in]{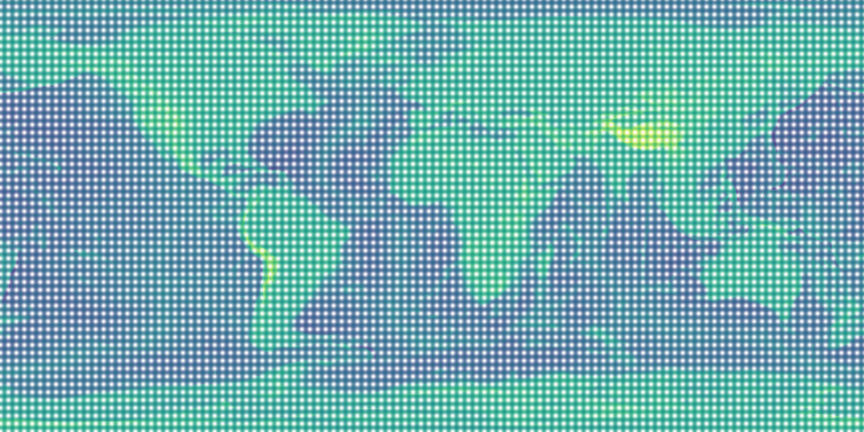}%
    \label{fig:bathymetry2}}
    \hfil
         \subfloat[$m=2.33$M, $N=37.3$M.]{\includegraphics[width=2in]{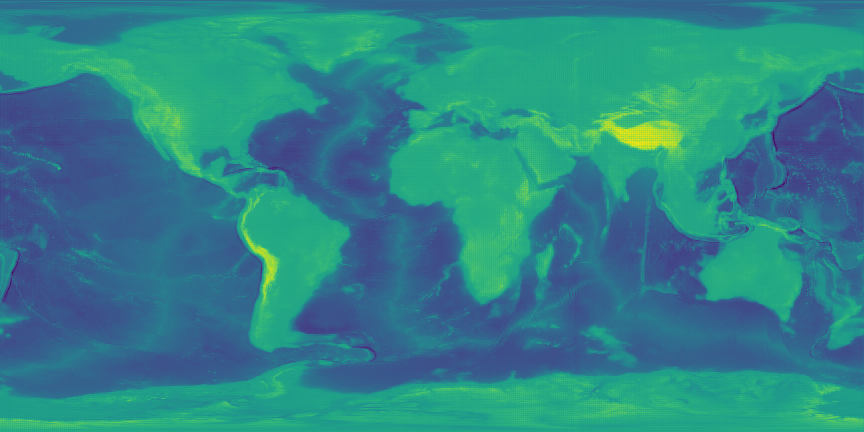}%
    \label{fig:bathymetry3}}
        \caption{Bathymetry dataset reconstruction with GP regression. The color corresponds to the posterior predicted elevation of the earth surface both above and below the sea, and the opacity is inversely proportional to the variance of the approximate GP prediction in each location.}
        \label{fig:Bathymetry_predictions}
\end{figure*}

\subsection{Mapping daily precipitation levels}\label{sec:precip_results}

In this section, we compare the prediction accuracy and computation time of our mapping approach to alternative methods on a large geo-spatial dataset which is used as a benchmark dataset for evaluation accuracy and computation time by~\cite{yadav_faster_2021}, and~\cite{pmlr-v37-wilson15}. The dataset contains $528\:474$ measurements of daily precipitations from the US in the training set, and $100\:000$ measurements in the test set~\cite{yadav_faster_2021}. The input dimensions are latitude, longitude, and time. We use the squared exponential kernel using the same hyperparameters as~\cite{yadav_faster_2021}. These hyperparameters are $\sigma_{\text{SE}}=3.99$, $\sigma_{\text{y}}=2.789$ and $l_{\text{SE}}=[3.094, 2.030, 0.189]$, where the three lengthscales $l_{\text{SE}}$ apply to each of the three dimensions, respectively.

In Table~\ref{tab:SMSEs}, the standardized mean squared error (SMSE) of our approach with $r^{\star}=3.5l_{\text{SE}}$ is compared to the full GP prediction, and to the inducing input approximation with basis functions placed on the same grid. The SMSE of our approach almost matches the inducing input approximation with the same number of basis functions. Using $200~000$ basis functions, it matches the GP prediction accuracy with the same number of measurements. Using all the measurements and $200~000$ basis functions gives the highest prediction accuracy, which is a combination that is computationally infeasible given our hardware constraints for both the full GP regression and the inducing input approximation to give a prediction. However, our proposed method has an online training time of only $0.017$ seconds per measurement and $0.016$ seconds per prediction.

\subsection{Global Bathymetry Field Mapping}\label{sec:bathymetry_data}

To investigate the time required for our method to include a new measurement and make a prediction in a large geospatial field with fine-scale variations, we run it on a dataset containing values of the height difference with respect to sea level across the globe \cite{gebco_compilation_group_gridded_2021}. The input domain of this data is huge compared to the scale of the spatial variations, and the data is therefore challenging to train on using state-of-the-art methods. We retrieve $37.5$ million depth values from the database, and we consider the latitude and longitude as input locations.

We test our approach using the squared exponential kernel from~\eqref{eq:SE_kernel}. We choose $\sigma_{\text{SE}}^2$ equal to the variance of the $37.5$ million measurements and $\sigma_{\text{y}}=0.1\sigma_{\text{SE}}$ to ensure a reasonable signal-to-noise ratio. Furthermore, we set the lengthscale $l_{\text{SE}}=0.16$ degrees, which corresponds to $18.3$ km on the equator. Note that this lengthscale can straightforwardly be changed to a more physically informed value by an end-user. 

The result in Fig.~\ref{fig:Bathymetry_predictions} shows the bathymetry map learned using $10\%$ of the measurements and $2.33$ million (M) basis functions (Fig.~\ref{fig:bathymetry1}), the map learned using $100\%$ of the measurements and $268$ thousand (K) basis functions (Fig.~\ref{fig:bathymetry2}), and map learned using $100\%$ of the measurements and a dense grid of $2.33$ million basis functions (Fig.~\ref{fig:bathymetry3}). We use a local subset contained within $r^{\star}=3l_{\text{SE}}$ to approximate the GP posterior mean. For the results in Fig.~\ref{fig:bathymetry3}, including each new measurement takes $3.7\times10^{-4}\pm0.12\times10^{-4}$ seconds. Each prediction takes $0.0097\pm0.017$ seconds. These results demonstrate that our proposed approach can remain computationally feasible in cases where the measurement density is high, and the map area is large relative to the length scale of the spatial variations. The disadvantage of discarding measurements is visible in Fig.~\ref{fig:bathymetry1}. When fewer measurements are included, less information is available about the map, causing the image to appear blurry. The disadvantage of discarding basis functions is visible in Fig.~\ref{fig:bathymetry2}. When there are not enough basis functions, the map is not represented at a high enough resolution, causing the image to appear grid-like.

\begin{figure*}[]
\captionsetup[subfigure]{font=footnotesize}
\centering
\subfloat[Odometry]{\includegraphics[width=0.9in]{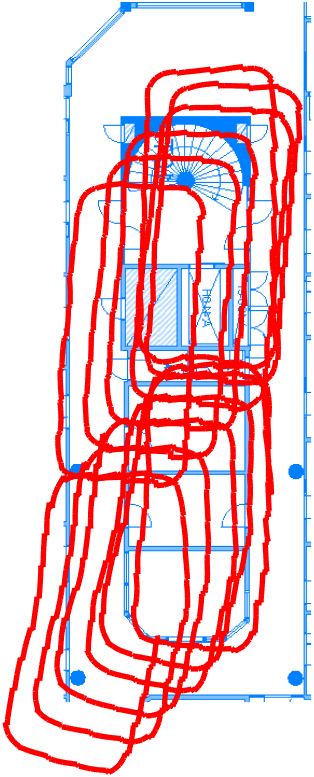}%
\label{fig:foot_odometry}}
\hfil
\subfloat[Our approach, $r=1.5l_{SE},\newline 2.8\pm 0.85 \text{ ms}$]{\includegraphics[width=0.9in]{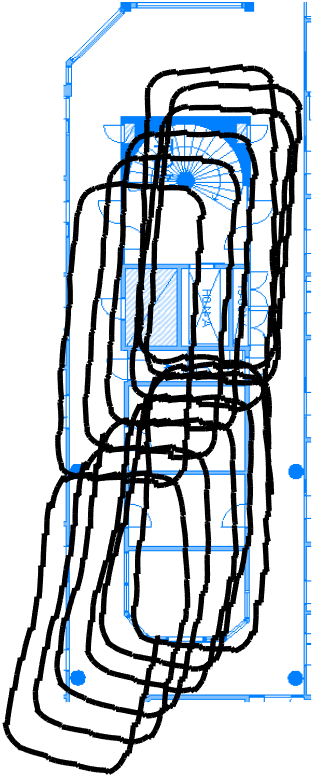}%
\label{fig:foot_r15}}
\hfil
\subfloat[Our approach, $r=2l_{SE}\newline 4.2\pm 1.1 \text{ ms}$]{\includegraphics[width=0.9in]{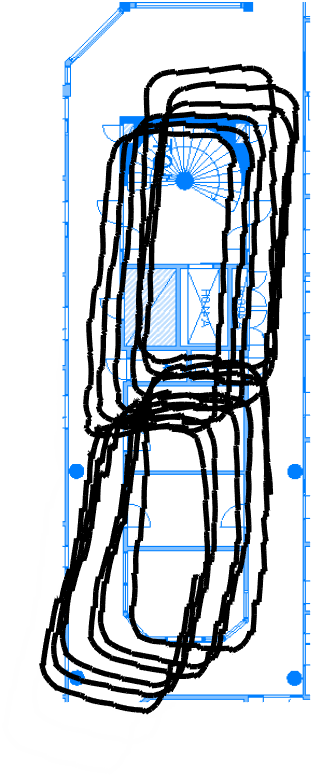}%
\label{fig:foot_r20}}
\hfil
\subfloat[Our approach, $r=2.5l_{SE}\newline 5.6\pm 1.3 \text{ ms}$]{\includegraphics[width=0.9in]{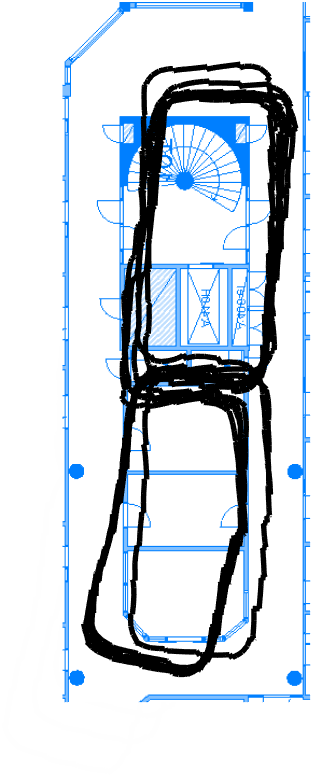}%
\label{fig:foot_r25}}
\hfil
\subfloat[Baseline, \newline $26\pm 4.6 \text{ ms}$]{\includegraphics[width=0.9in]{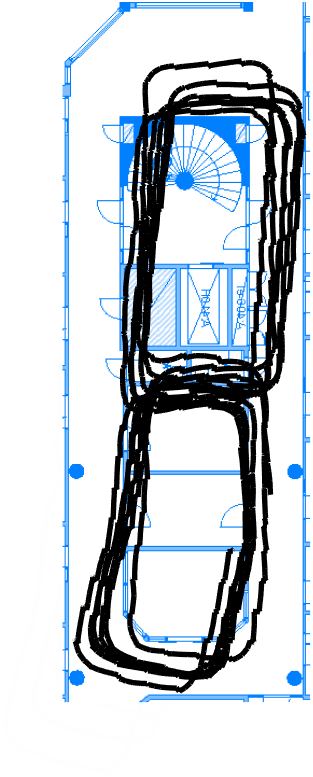}%
\label{fig:foot_hilbert}}
\caption{(a) Trajectory estimates for indoor pedestrian walking laps in a hallway using only foot-mounted odometry, (b--d) EKF Mag-SLAM with various sizes of the local domain determined by $r$, and (e) the baseline algorithm that uses with Hilbert space basis functions. The average computation time in milliseconds for one iteration of each filter (dynamic update + measurement update) is written below each subfigure.}
\label{fig_foot}
\end{figure*}

\subsection{Using local information filter for faster Mag-SLAM}

To experimentally compare our EKF Mag-SLAM approach from Section~\ref{sec:SLAM_information_form} with that from~\cite{viset_extended_2022}, we apply both algorithms to a dataset from a foot-mounted sensor that was collected by~\cite{skog_openshoe_2012} and subsequently used in~\cite{viset_extended_2022} to demonstrate the performance of the method. We measure the average computation time required for each iteration of our algorithm and compare this with the average computation time required, on the same laptop, for each iteration of the algorithm from~\cite{viset_extended_2022}. 

In Fig.~\ref{fig:foot_odometry}, the odometry obtained by using only accelerometer and gyroscope measurements from a foot-mounted sensor using the algorithm in~\cite{skog_inertial_2016} is displayed. In Fig.~\ref{fig:foot_hilbert}, the estimated trajectory using the EKF with Hilbert space basis functions as in~\cite{viset_extended_2022} is displayed. All estimates are overlayed on the floorplan of the building where the measurements were collected. This gives a rudimentary means of evaluating the position estimation accuracy, since the subject walked through the same hallways in a repeated pattern 8-motion. The drift in the odometry in Fig.~\ref{fig:foot_odometry} therefore shows up as a slow displacement of the position estimate away from the hallway. In contrast, the estimates obtained using our approximate mapping (see Fig.~\ref{fig:foot_r25}) and using the algorithm from~\cite{viset_extended_2022} (see  Fig.~\ref{fig:foot_hilbert}) compensate for this drift. The notable difference between our approach and the approach from~\cite{viset_extended_2022}, is that the latter uses Hilbert space basis functions resulting in a time of $26\pm 4.6$ ms to run at each iteration, while our equally accurate algorithm requires only $5.6\pm 1.3$ ms to run at each iteration. Although comparing the computational time of different algorithms depends heavily on implementation, we used the same implementation as~\cite{viset_extended_2022}, and wrote the implementation of our algorithm in the same language and on the same format, simply swapping the terms used by~\cite{viset_extended_2022} with ours. 

\section{Conclusion}\label{sec:conclusion}

To improve position estimation online with few computational resources, we have presented an efficient online mapping technique that approximates the GP posterior. The required number of computations neither scales with the number of measurements, nor with the spatial extent of the map. The storage requirements of our presented mapping algorithm scale linearly with the spatial extent of the map, and also does not scale with the number of measurements. We have also demonstrated the ability of our mapping algorithm to match the accuracy of previously proposed approximations on benchmark datasets using a lower computation time. Our mapping algorithm can also be used for SLAM. We have shown experimentally that our proposed method achieves the same prediction accuracy for magnetic field SLAM using a foot-mounted sensor as in previous work, while requiring a shorter computation time. Future work could investigate the existence of theoretical bounds on the approximation errors, or investigate ways to incorporate the mapping technique for SLAM in different geospatial fields, and for other applications such as navigation of robots or vehicles.

\appendices

\section{Prior covariance for the prediction-point dependent basis function approximation}\label{sec:P_star_derivation}

The prior for the parametric approximation to the GP regression using the local subset of basis functions is given by
\begin{equation}\label{eq:local_domain_prior}
    \tilde{f}^{\star}=\Phi_{\mathcal{S}^\star}^\top w_{\mathcal{S}^{\star}},\qquad w_{\mathcal{S}^{\star}}\sim\mathcal{N}(0,P_{\star}),
\end{equation}
where $P_{\star}$ is the prior covariance on the local weights. The condition that the prior should be recovered in the center locations of the basis functions is given as
\begin{equation}\label{eq:criteria}
    p(\tilde{f}^\star(u_{S^{\star}}))=p(f(u_{S^{\star}}))),
\end{equation}
where $u_{S^{\star}}$ denotes the center locations of the basis functions contained in the set $S^{\star}$. As both distributions in~\eqref{eq:criteria} are normal distributions with mean $0$, this condition holds if and only if the two covariances are equal according to
\begin{equation}\label{eq:Phi_star_general}
    \Phi_{S^\star}(u_{S^\star})^\tp P_{\star} \Phi_{S^\star}(u_{S^\star}) = K(u_{S^\star},u_{S^\star}).
\end{equation}
This results in the closed-form expression for $P_{\star}$
\begin{equation}
    P_{\star}=(\Phi_{S^\star}(u_{S^\star})^\tp)^{-1} K(u_{S^\star},u_{S^\star}) \Phi_{S^\star}(u_{S^\star})^{-1}, 
\end{equation}
when $(\Phi_{S^\star}(u_{S^\star})^\tp)$ and $\Phi_{S^\star}(u_{S^\star})$ are invertible. The entry in row $i$ and column $j$ of the matrix $\Phi_{S^\star}(u_{S^\star})$ is $\phi_i(u_j)$, which is defined using \eqref{eq:finite_support_inducing_functions} as
\begin{equation}
    \phi_i(u_j)=\left\{\begin{matrix}
\kappa(u_i,u_j), & \|u_i-u_j \|_{\infty}\leq r \\ 
0, & \|u_i-u_j \|_{\infty}> r
\end{matrix}\right.
\end{equation}
For all inducing input pairs $i,j$ in the set $\mathcal{S}(x^{\star},r^{\star})$, it holds that $\|u_i-u_j\|_{\infty}\leq 2 r^{\star}$. If $r\geq2r^{\star}$, the condition $\|u_i-u_j \|_{\infty}\leq r $ holds for all $i,j \in\mathcal{S}^{\star}$. This implies that $\phi_i(u_j)=\kappa(u_i,u_j)$ for all $i,j \in \mathcal{S}^{\star}$ and hence $\Phi_{S^\star}(u_{S^\star})=K(u_{S^\star},u_{S^\star})$. Inserting this result into~\eqref{eq:Phi_star_general} gives $P_{\star}=K(u_{S^\star},u_{S^\star})^{-1}$. 

\bibliography{bibliography2} 

\begin{thebibliography}{10}
\providecommand{\url}[1]{#1}
\csname url@samestyle\endcsname
\providecommand{\newblock}{\relax}
\providecommand{\bibinfo}[2]{#2}
\providecommand{\BIBentrySTDinterwordspacing}{\spaceskip=0pt\relax}
\providecommand{\BIBentryALTinterwordstretchfactor}{4}
\providecommand{\BIBentryALTinterwordspacing}{\spaceskip=\fontdimen2\font plus
\BIBentryALTinterwordstretchfactor\fontdimen3\font minus \fontdimen4\font\relax}
\providecommand{\BIBforeignlanguage}[2]{{%
\expandafter\ifx\csname l@#1\endcsname\relax
\typeout{** WARNING: IEEEtran.bst: No hyphenation pattern has been}%
\typeout{** loaded for the language `#1'. Using the pattern for}%
\typeout{** the default language instead.}%
\else
\language=\csname l@#1\endcsname
\fi
#2}}
\providecommand{\BIBdecl}{\relax}
\BIBdecl

\bibitem{gustafsson_statistical_2013}
F.~Gustafsson, \emph{Statistical {Sensor} {Fusion}}.\hskip 1em plus 0.5em minus 0.4em\relax Studentliteratur, 2013.

\bibitem{woodman_introduction_2007}
\BIBentryALTinterwordspacing
O.~J. Woodman, ``An introduction to inertial navigation,'' 2007. [Online]. Available: \url{https://www.cl.cam.ac.uk/techreports/UCAM-CL-TR-696.pdf}
\BIBentrySTDinterwordspacing

\bibitem{kok_using_2017-1}
M.~Kok, J.~Hol, and T.~B. Sch{\"o}n, ``Using {{Inertial Sensors}} for {{Position}} and {{Orientation Estimation}},'' \emph{Foundations and Trends on Signal Processing}, Jan. 2017.

\bibitem{rasmussen_gaussian_2005}
C.~E. Rasmussen and C.~K.~I. Williams, \emph{Gaussian {Processes} for {Machine} {Learning}}.\hskip 1em plus 0.5em minus 0.4em\relax The MIT Press, Nov. 2005.

\bibitem{wahlstrom_modeling_2015}
N.~Wahlström, ``Modeling of {Magnetic} {Fields} and {Extended} {Object} for {Localization} {Applications},'' {PhD} {Thesis}, Link\"oping University, Dec. 2015.

\bibitem{baileyD:2006}
T.~Bailey and H.~Durrant-Whyte, ``Simultaneous localization and mapping {(SLAM)}: {Part II},'' \emph{IEEE Robotics \& Automation Magazine}, vol.~13, no.~3, pp. 108--117, 2006.

\bibitem{durrantWhyteB:2006}
H.~Durrant-Whyte and T.~Bailey, ``Simultaneous localization and mapping: {P}art~{I},'' \emph{IEEE Robotics \& Automation Magazine}, vol.~13, no.~2, pp. 99--110, 2006.

\bibitem{kok_scalable_2018}
M.~Kok and A.~Solin, ``Scalable {Magnetic} {Field} {SLAM} in {3D} {Using} {Gaussian} {Process} {Maps},'' in \emph{proceedings of the 21st {International} {Conference} on {Information} {Fusion} ({FUSION})}, Jul. 2018, pp. 1353--1360.

\bibitem{robertson_simultaneous_2013}
P.~Robertson, M.~Frassl, M.~Angermann, M.~Doniec, B.~J. Julian, M.~Garcia~Puyol, M.~Khider, M.~Lichtenstern, and L.~Bruno, ``Simultaneous {Localization} and {Mapping} for pedestrians using distortions of the local magnetic field intensity in large indoor environments,'' in \emph{proceedings of the {International} {Conference} on {Indoor} {Positioning} and {Indoor} {Navigation}}, Oct. 2013, pp. 1--10.

\bibitem{jung_indoor_2015}
J.~Jung, S.-M. Lee, and H.~Myung, ``Indoor {Mobile} {Robot} {Localization} and {Mapping} {Based} on {Ambient} {Magnetic} {Fields} and {Aiding} {Radio} {Sources},'' \emph{IEEE Transactions on Instrumentation and Measurement}, vol.~64, no.~7, pp. 1922--1934, Jul. 2015.

\bibitem{vallivaara_magnetic_2011}
I.~Vallivaara, J.~Haverinen, A.~Kemppainen, and J.~Röning, ``Magnetic field-based {SLAM} method for solving the localization problem in mobile robot floor-cleaning task,'' in \emph{proceedings of the 15th {International} {Conference} on {Advanced} {Robotics} ({ICAR})}, Jun. 2011, pp. 198--203.

\bibitem{torroba_online_2023}
I.~Torroba, M.~Cella, A.~Terán, N.~Rolleberg, and J.~Folkesson, ``Online {Stochastic} {Variational} {Gaussian} {Process} {Mapping} for {Large}-{Scale} {Bathymetric} {SLAM} in {Real} {Time},'' \emph{IEEE Robotics and Automation Letters}, vol.~8, no.~6, pp. 3150--3157, Jun. 2023.

\bibitem{barkby_bathymetric_2011}
S.~Barkby, S.~B. Williams, O.~Pizarro, and M.~V. Jakuba, ``Bathymetric {SLAM} with no map overlap using {Gaussian} {Processes},'' \emph{proceedings of the International Conference on Intelligent Robots and Systems}, pp. 1242--1248, Sep. 2011.

\bibitem{kjaergaard_terrain_2011}
M.~Kjaergaard, E.~Bayramoglu, A.~S. Massaro, and K.~Jensen, ``Terrain {Mapping} and {Obstacle} {Detection} {Using} {Gaussian} {Processes},'' in \emph{proceedings of the 10th {International} {Conference} on {Machine} {Learning} and {Applications} and {Workshops}}, vol.~1, Dec. 2011, pp. 118--123.

\bibitem{yu_terrain_2018}
H.~Yu and B.~Lee, ``Terrain field {SLAM} and {Uncertainty} {Mapping} using {Gaussian} {Process},'' in \emph{Proceedings of the 18th {International} {Conference} on {Control}, {Automation} and {Systems}}, Oct. 2018, pp. 1077--1080.

\bibitem{solin_hilbert_2014}
A.~Solin and S.~Särkkä, ``Hilbert space methods for reduced-rank {Gaussian} process regression,'' \emph{Statistics and Computing}, vol.~30, pp. 419--446, 2014.

\bibitem{jung_indoor_2014}
J.~Jung, S.-M. Lee, and H.~Myung, ``\BIBforeignlanguage{en}{Indoor {Mobile} {Robot} {Localization} {Using} {Ambient} {Magnetic} {Fields} and {Range} {Measurements}},'' in \emph{\BIBforeignlanguage{en}{Robot {Intelligence} {Technology} and {Applications} 2: {Results} from the 2nd {International} {Conference} on {Robot} {Intelligence} {Technology} and {Applications}}}, ser. Advances in {Intelligent} {Systems} and {Computing}, J.-H. Kim, E.~T. Matson, H.~Myung, P.~Xu, and F.~Karray, Eds.\hskip 1em plus 0.5em minus 0.4em\relax Cham: Springer International Publishing, 2014, pp. 137--143.

\bibitem{viseras_decentralized_2016}
A.~Viseras, T.~Wiedemann, C.~Manss, L.~Magel, J.~Mueller, D.~Shutin, and L.~Merino, ``Decentralized multi-agent exploration with online-learning of {Gaussian} processes,'' in \emph{Proceedings of the {International} {Conference} on {Robotics} and {Automation}}, May 2016, pp. 4222--4229.

\bibitem{ding_multiresolution_2017}
Y.~Ding, R.~Kondor, and J.~Eskreis-Winkler, ``Multiresolution {Kernel} {Approximation} for {Gaussian} {Process} {Regression},'' in \emph{Advances in {Neural} {Information} {Processing} {Systems}}, vol.~30.\hskip 1em plus 0.5em minus 0.4em\relax Curran Associates, Inc., 2017.

\bibitem{viset_extended_2022}
F.~Viset, R.~Helmons, and M.~Kok, ``\BIBforeignlanguage{en}{An {Extended} {Kalman} {Filter} for {Magnetic} {Field} {SLAM} {Using} {Gaussian} {Process} {Regression}},'' \emph{\BIBforeignlanguage{en}{Sensors}}, vol.~22, no.~8, p. 2833, Jan. 2022.

\bibitem{jang_multi-robot_2020}
D.~Jang, J.~Yoo, C.~Y. Son, D.~Kim, and H.~J. Kim, ``Multi-{Robot} {Active} {Sensing} and {Environmental} {Model} {Learning} {With} {Distributed} {Gaussian} {Process},'' \emph{IEEE Robotics and Automation Letters}, vol.~5, no.~4, pp. 5905--5912, Oct. 2020.

\bibitem{viset_magnetic_2021}
F.~Viset, J.~T. Gravdahl, and M.~Kok, ``Magnetic field norm {SLAM} using {Gaussian} process regression in foot-mounted sensors,'' in \emph{Proceedings of the {European} {Control} {Conference} ({ECC})}, Jun. 2021, pp. 392--398.

\bibitem{bijl_online_2015}
H.~Bijl, J.-W. van Wingerden, T.~B.~Schön, and M.~Verhaegen, ``\BIBforeignlanguage{en}{Online sparse {Gaussian} process regression using {FITC} and {PITC} approximations},'' \emph{\BIBforeignlanguage{en}{IFAC-PapersOnLine}}, vol.~48, no.~28, pp. 703--708, Jan. 2015.

\bibitem{quinonero-candela_unifying_2005}
J.~Quiñonero-Candela and C.~E. Rasmussen, ``A {Unifying} {View} of {Sparse} {Approximate} {Gaussian} {Process} {Regression},'' \emph{The Journal of Machine Learning Research}, vol.~6, pp. 1939--1959, Dec. 2005.

\bibitem{hensman_variational_2016}
J.~Hensman, N.~Durrande, and A.~Solin, ``Variational {Fourier} features for {Gaussian} processes,'' \emph{Journal of Machine Learning Research}, vol.~18, Nov. 2016.

\bibitem{viset:2024}
F.~M. Viset, ``Scalable magnetic field mapping and localization using gaussian process regression,'' {PhD} {Thesis}, Delft University of Technology, Nov. 2024.

\bibitem{kullberg_online_2021}
A.~Kullberg, I.~Skog, and G.~Hendeby, ``Online {Joint} {State} {Inference} and {Learning} of {Partially} {Unknown} {State}-{Space} {Models},'' \emph{IEEE Transactions on Signal Processing}, vol.~69, pp. 4149--4161, 2021.

\bibitem{julier:2001}
S.~J. Julier, ``A sparse weight {K}alman filter approach to simultaneous localisation and map building,'' in \emph{Proceedings of the IEEE/RSJ International Conference on Intelligent Robots and Systems. Expanding the Societal Role of Robotics in the the Next Millennium (Cat. No. 01CH37180)}, vol.~3, 2001, pp. 1251--1256.

\bibitem{karlsson_bayesian_2006}
R.~Karlsson and F.~Gustafsson, ``Bayesian {Surface} and {Underwater} {Navigation},'' \emph{IEEE Transactions on Signal Processing}, vol.~54, no.~11, pp. 4204--4213, Nov. 2006.

\bibitem{vasudevan_gaussian_2009}
S.~Vasudevan, F.~Ramos, E.~Nettleton, H.~Durrant-Whyte, and A.~Blair, ``Gaussian {Process} modeling of large scale terrain,'' in \emph{proceedings of {International} {Conference} on {Robotics} and {Automation}}, May 2009, pp. 1047--1053, iSSN: 1050-4729.

\bibitem{tichavsky_grid-based_2023}
P.~Tichavský, O.~Straka, and J.~Duník, ``Grid-{Based} {Bayesian} {Filters} {With} {Functional} {Decomposition} of {Transient} {Density},'' \emph{IEEE Transactions on Signal Processing}, vol.~71, pp. 92--104, 2023.

\bibitem{gustafsson_particle_2002}
F.~Gustafsson, F.~Gunnarsson, N.~Bergman, U.~Forssell, J.~Jansson, R.~Karlsson, and P.-J. Nordlund, ``Particle filters for positioning, navigation, and tracking,'' \emph{IEEE Transactions on Signal Processing}, vol.~50, no.~2, pp. 425--437, 2002.

\bibitem{osman_indoor_2022}
M.~Osman, F.~Viset, and M.~Kok, ``Indoor {SLAM} using a foot-mounted {IMU} and the local magnetic field,'' in \emph{Proceedings of the 25th {International} {Conference} on {Information} {Fusion}}, Jul. 2022, pp. 1--7.

\bibitem{gramacy_local_2015}
R.~B. Gramacy and D.~W. Apley, ``Local {Gaussian} {Process} {Approximation} for {Large} {Computer} {Experiments},'' \emph{Journal of Computational and Graphical Statistics}, vol.~24, no.~2, pp. 561--578, 2015.

\bibitem{snelson_local_2007-1}
E.~Snelson and Z.~Ghahramani, ``\BIBforeignlanguage{en}{Local and global sparse {Gaussian} process approximations},'' in \emph{\BIBforeignlanguage{en}{Proceedings of the {Eleventh} {International} {Conference} on {Artificial} {Intelligence} and {Statistics}}}.\hskip 1em plus 0.5em minus 0.4em\relax PMLR, Mar. 2007, pp. 524--531, iSSN: 1938-7228.

\bibitem{park_efficient_2016}
C.~Park and J.~Z. Huang, ``Efficient {Computation} of {Gaussian} {Process} {Regression} for {Large} {Spatial} {Data} {Sets} by {Patching} {Local} {Gaussian} {Processes},'' \emph{Journal of Machine Learning Research}, vol.~17, no. 174, pp. 1--29, 2016.

\bibitem{park_domain_2011}
C.~Park, J.~Z. Huang, and Y.~Ding, ``Domain {Decomposition} {Approach} for {Fast} {Gaussian} {Process} {Regression} of {Large} {Spatial} {Data} {Sets},'' \emph{Journal of Machine Learning Research}, vol.~12, no.~47, pp. 1697--1728, 2011.

\bibitem{pmlr-v37-wilson15}
A.~Wilson and H.~Nickisch, ``Kernel interpolation for scalable structured gaussian processes ({KISS}-{GP}),'' in \emph{Proceedings of the 32nd {International} {Conference} on {Machine} {Learning}}, vol.~37, Jul. 2015, pp. 1775--1784.

\bibitem{yadav_faster_2021}
M.~Yadav, D.~Sheldon, and C.~Musco, ``\BIBforeignlanguage{en}{Faster {Kernel} {Interpolation} for {Gaussian} {Processes}},'' in \emph{\BIBforeignlanguage{en}{Proceedings of {the} 24th {International} {Conference} on {Artificial} {Intelligence} and {Statistics}}}, Mar. 2021, pp. 2971--2979.

\bibitem{mutambara_decentralized_1998}
A.~G.~O. Mutambara, \emph{\BIBforeignlanguage{en}{Decentralized {Estimation} and {Control} for {Multisensor} {Systems}}}.\hskip 1em plus 0.5em minus 0.4em\relax CRC Press, Jan. 1998.

\bibitem{wahlstrom_modeling_2013}
N.~Wahlström, M.~Kok, T.~B. Schön, and F.~Gustafsson, ``Modeling magnetic fields using {Gaussian} processes,'' in \emph{Proceedings of the {International} {Conference} on {Acoustics}, {Speech} and {Signal} {Processing}}, May 2013, pp. 3522--3526.

\bibitem{walter_exactly_2007}
M.~R. Walter, R.~M. Eustice, and J.~J. Leonard, ``\BIBforeignlanguage{en}{Exactly {Sparse} {Extended} {Information} {Filters} for {Feature}-based {SLAM}},'' \emph{\BIBforeignlanguage{en}{The International Journal of Robotics Research}}, vol.~26, no.~4, pp. 335--359, Apr. 2007.

\bibitem{turner_statistical_2010}
R.~E. Turner, ``Statistical {Models} for {Natural} {Sounds},'' Ph.D. dissertation, Gatsby Computational Neuroscience Unit, UCL, 2010.

\bibitem{gebco_compilation_group_gridded_2021}
\BIBentryALTinterwordspacing
{GEBCO Compilation Group}, ``\BIBforeignlanguage{en}{Gridded bathymetry data {GEBCO} ({General} {Bathymetric} {Chart} of the {Oceans}) {Grid}.}'' 2021. [Online]. Available: \url{gebco.net/data_and_products/gridded_bathymetry_data/}
\BIBentrySTDinterwordspacing

\bibitem{skog_openshoe_2012}
I.~Skog, ``{OpenShoe} {Matlab} {Framework},'' 2012.

\bibitem{skog_inertial_2016}
I.~Skog, J.-O. Nilsson, P.~Händel, and A.~Nehorai, ``Inertial {Sensor} {Arrays}, {Maximum} {Likelihood}, and {Cram\'er}-{Rao} {Bound},'' \emph{IEEE Transactions on Signal Processing}, vol.~64, pp. 1--1, Aug. 2016.

\end{thebibliography}
\bibliographystyle{IEEEtran}



\begin{IEEEbiography}[{\includegraphics[width=1in,height=1.25in,clip,keepaspectratio]{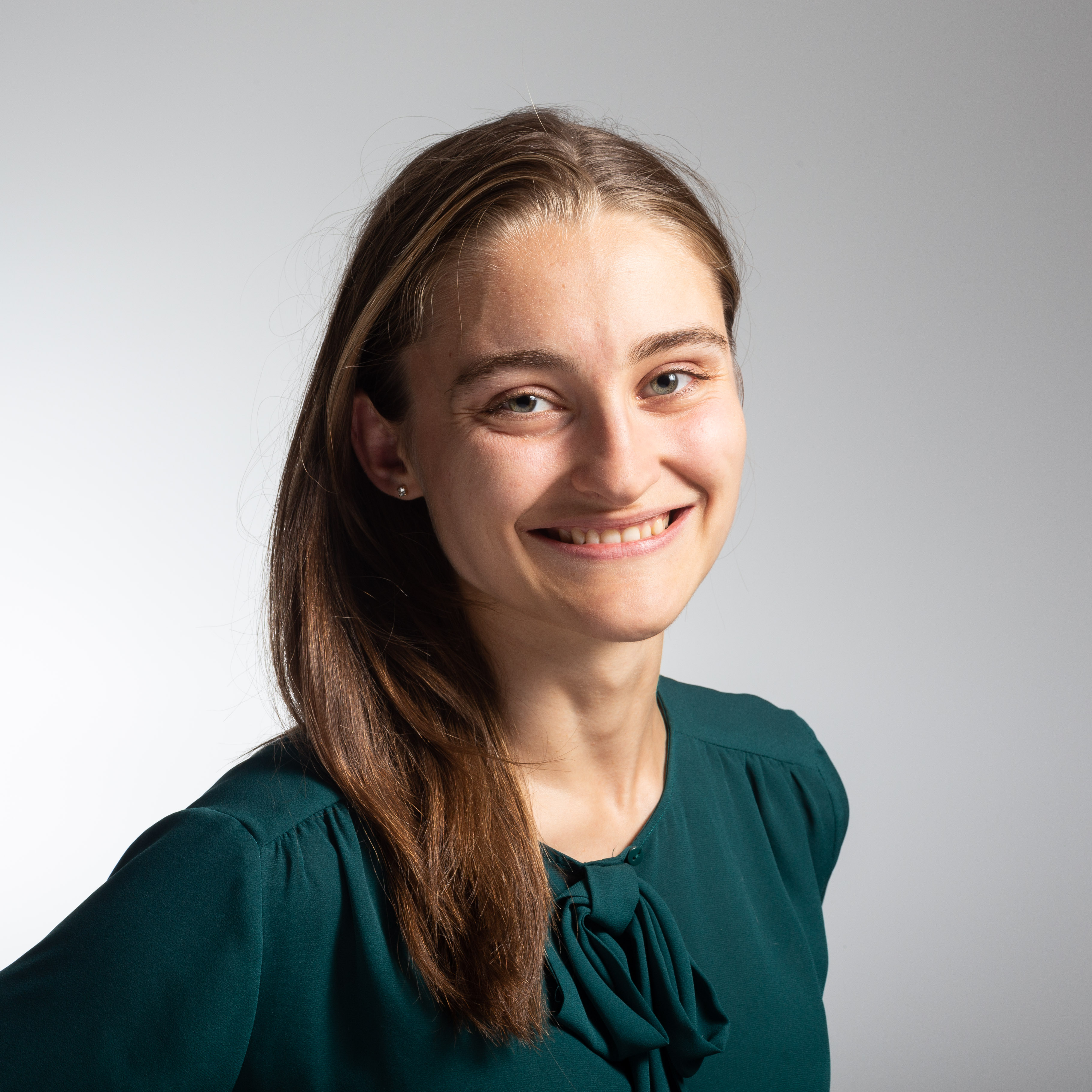}}]{Frida Viset}
 (1996) obtained her MSc in Cybernetics and Robotics in 2020 from the Norwegian University of Technology and her PhD in November 2024 from Delft University of Technology. Her research interests are within Kalman filters, scalability, and Gaussian process regression.
\end{IEEEbiography}

\begin{IEEEbiography}[{\includegraphics[width=1in,height=1.25in,clip,keepaspectratio]{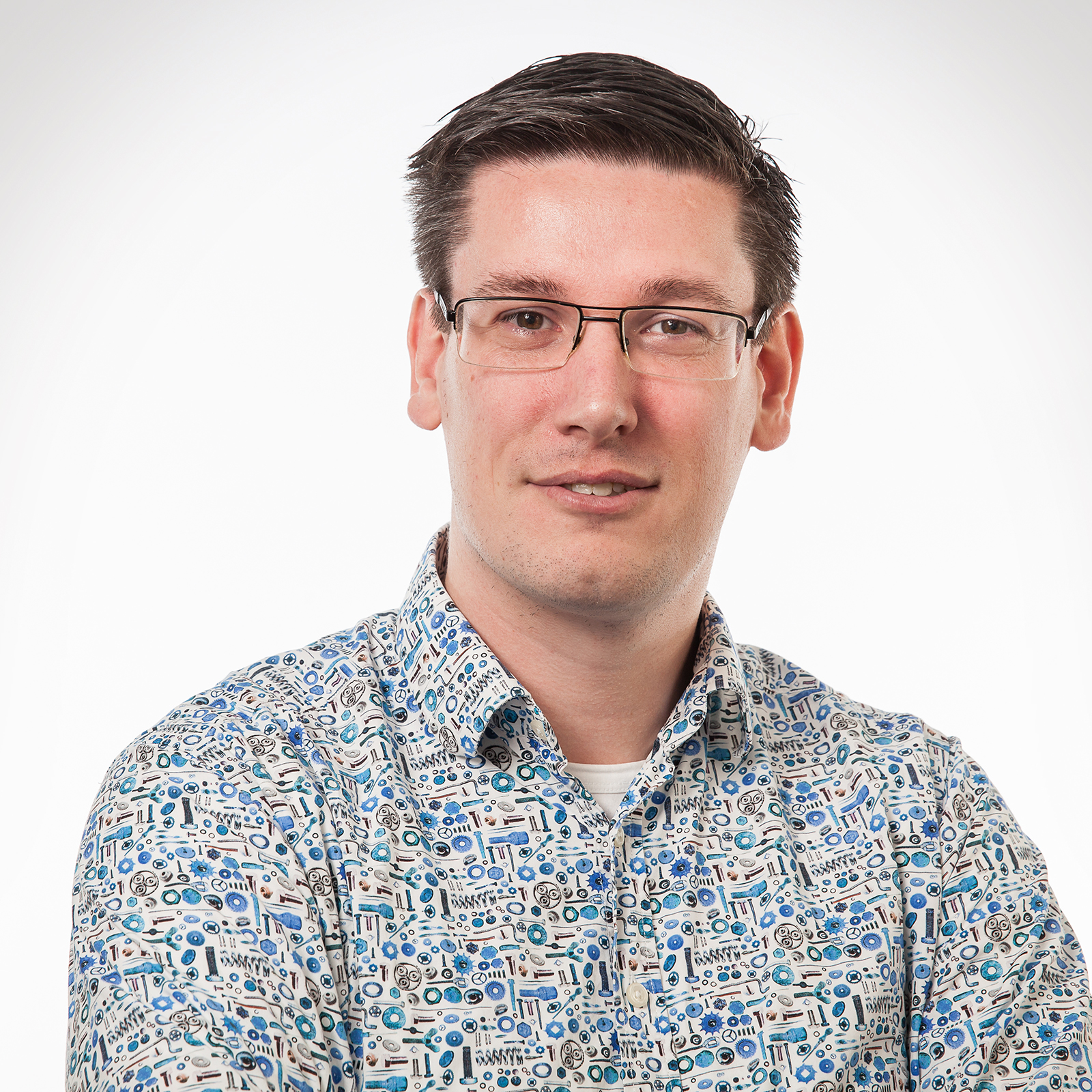}}]{Rudy Helmons}
 (1987) obtained his MSc in Mechanical Engineering in 2011 after which he started as a research engineer at Royal IHC. He obtained his PhD `cum laude' from TU Delft in 2017 on the topic of rock excavation in underwater conditions. After that, he started as an assistant professor in Offshore and Dredging Engineering at TU Delft, and as of February 2020, he is also Adjunct Associate Professor for deep sea mining at NTNU. His research interests are related to underwater automation of mining equipment, large scale terrain mapping, and enabling a transition towards cost-effective solutions to monitor the deep-sea environment at large temporal and spatial scales. 
\end{IEEEbiography}

\begin{IEEEbiography}[{\includegraphics[width=1in,height=1.25in,clip,keepaspectratio]{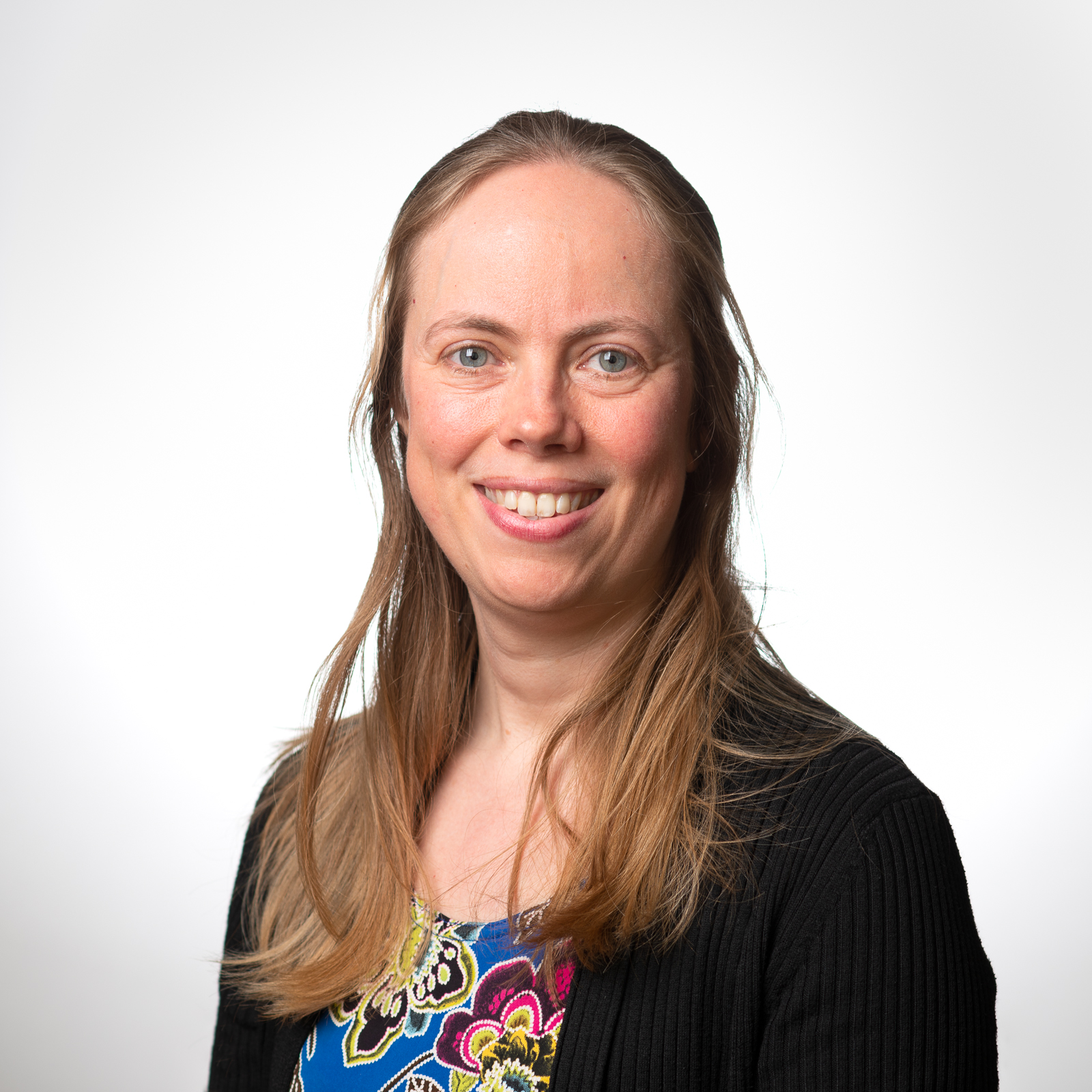}}]{Manon Kok}
 received the M.Sc. degrees in applied physics and in philosophy of science, technology and society from the University of Twente, Enschede, the Netherlands, in 2007 and 2009, respectively, and the Ph.D. degree in automatic control from Linköping University, Linköping, Sweden, in 2017. From 2009 to 2011, she was a Research Engineer with Xsens Technologies. From 2017 to 2018 she was a Postdoctoral Researcher with the Computational and Biological Learning Laboratory, Machine Learning Group, University of Cambridge, Cambridge, U.K. She is currently an Associate Professor with the Delft Center for Systems and Control, Delft University of Technology, the Netherlands. Her research interests include the fields of probabilistic inference for sensor fusion, signal processing, and machine learning.
\end{IEEEbiography}

\vspace{11pt}

\vfill

\end{document}